\documentclass[10pt,twocolumn,letterpaper]{article}   

\usepackage{cuted}
\usepackage[pagenumbers]{cvpr}
\usepackage{multirow}
\usepackage{bm}
\definecolor{cvprblue}{rgb}{0.21,0.49,0.74}
\usepackage[pagebackref,breaklinks,colorlinks,allcolors=cvprblue]{hyperref}
\usepackage{xcolor}
\usepackage{makecell}
\usepackage{marvosym}
\usepackage{booktabs} 
\usepackage{pdfpages}
\usepackage{tablefootnote}

\def\paperID{****}
\def\confName{CVPR}
\def\confYear{2025}

\title{OVO-Bench: How Far is Your Video-LLMs from Real-World Online Video Understanding?}

\author{\textbf{Yifei Li}$^{*1,2 \dag}$, \textbf{Junbo Niu}$^{*1,3 \dag}$,  \textbf{Ziyang Miao}$^{3}$, \textbf{Chunjiang Ge}$^{2}$, \textbf{Yuanhang Zhou}$^{2}$, \textbf{Qihao He}$^{4}$, \\ \textbf{Xiaoyi Dong}$^{1,5}${\textsuperscript{\Letter}}, \textbf{Haodong Duan}$^{1}$, \textbf{Shuangrui Ding}$^{1,5 \dag}$, \textbf{Rui Qian}$^{1,5 \dag}$, \\ \textbf{Pan Zhang}$^{1}$, \textbf{Yuhang Zang}$^{1}$, \textbf{Yuhang Cao}$^{1}$,   \textbf{Conghui He}$^{6}$, \textbf{Jiaqi Wang}$^{1,7}${\textsuperscript{\Letter}}\\
$^1$Shanghai Artificial Intelligence Laboratory,  $^2$Tsinghua University, \\ $^3$ Beihang University, $^4$ Communication University of China,  \\ $^5$ The Chinese University of Hong Kong, $^6$ SenseTime Group, \\ $^7$ Shanghai Innovation Institute
}

\begin{document}
\maketitle
{\let\thefootnote\relax\footnotetext{\noindent* indicates equal contribution. $\dag$ indicates interns at IXCLab, Shanghai AI Laboratory}}

\begin{strip}
    \centering
    \includegraphics[width=0.65\textwidth]{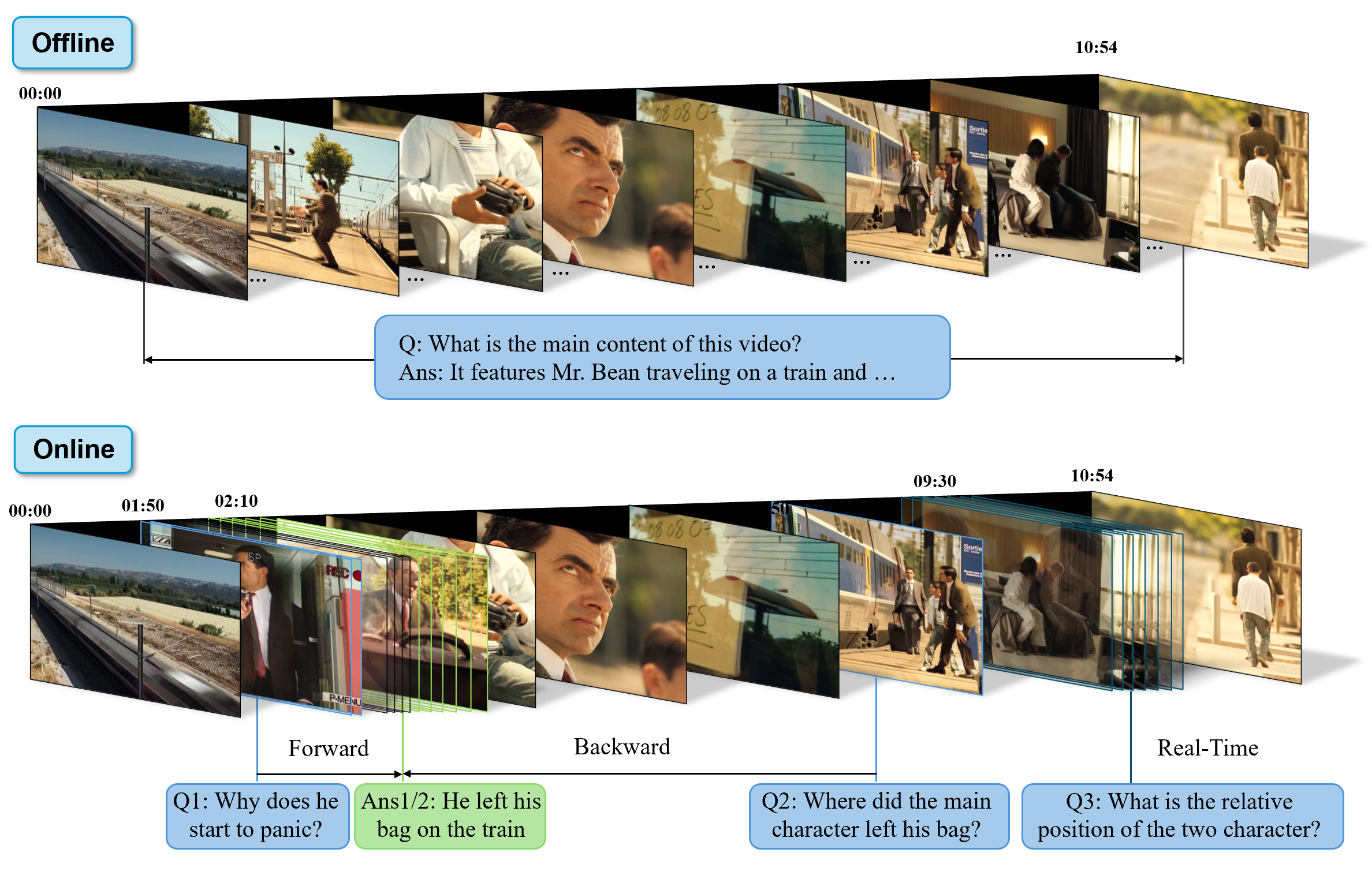} 
    \vspace{-1em}
    \captionof{figure}{
    A demonstrative comparison between offline and online video understanding~\cite{videollm-online}. 
    Offline video understanding focuses on answering questions based on the entirety of a video. In contrast, online video understanding involves posing queries about the context of a video at intermediate points, demanding the ability to trace back past information, perceive ongoing events, and adapt to continuous input.
    } \label{fig:teaser}
\end{strip}
\vspace{-20pt}

\begin{abstract}

\noindent

Temporal Awareness—the ability to reason dynamically based on the timestamp when a question is raised—is the key distinction between offline and online video LLMs. Unlike offline models, which rely on complete videos for static, post hoc analysis, online models process video streams incrementally and dynamically adapt their responses based on the timestamp at which the question is posed. Despite its significance, temporal awareness has not been adequately evaluated in existing benchmarks.
To fill this gap, we present \textbf{OVO-Bench} (\textbf{O}nline-\textbf{V}ide\textbf{O}-\textbf{Bench}mark), a novel video benchmark that emphasizes the importance of timestamps for advanced online video understanding capability benchmarking. OVO-Bench evaluates the ability of video LLMs to reason and respond to events occurring at specific timestamps under three distinct scenarios: (1) \textbf{Backward tracing}: trace back to past events to answer the question. (2) \textbf{Real-time understanding}: understand and respond to events as they unfold at the current timestamp. (3) \textbf{Forward active responding}: delay the response until sufficient future information becomes available to answer the question accurately.
OVO-Bench comprises 12 tasks, featuring 644 unique videos and approximately human-curated 2,800 fine-grained meta-annotations with precise timestamps. We combine automated generation pipelines with human curation. With these high-quality samples, we further developed an evaluation pipeline to systematically query video LLMs along the video timeline.
Evaluations of eleven Video-LLMs reveal that, despite advancements on traditional benchmarks, current models struggle with online video understanding, showing a significant gap compared to human agents. We hope OVO-Bench will drive progress in video LLMs and inspire future research in online video reasoning. Our benchmark and code can be accessed at \href{https://github.com/JoeLeelyf/OVO-Bench}{https://github.com/JoeLeelyf/OVO-Bench}.

\end{abstract}    
\section{Introduction}
\label{sec:intro}

Large Vision Language Models (LVLMs)~\cite{Qwen2VL,liu2023llava,zhang2023internlmxcomposer,openai2023gpt4}
and Video-LLMs~\cite{damonlpsg2023videollama,li2023videochat,Maaz2023VideoChatGPT}
have shown remarkable progress, achieving impressive scores on existing benchmarks~\cite{fang2024mmbench,fu2024video,li2024mvbench}. Recent works, such as VideoLLM-online~\cite{videollm-online} and Flash-VStream~\cite{zhang2024flashvstreammemorybasedrealtimeunderstanding}, have pioneered J.A.R.V.I.S\footnote{J.A.R.V.I.S. is a fictional AI assistant from Marvel's Iron Man and Avengers series.}-like real-world video assistants by integrating pre-trained vision encoders~\cite{radford2019language} with LLMs~\cite{chiang2023vicuna,touvron2023llama}.
However, a critical question remains: \textit{How far are current state-of-the-art models from achieving human-level online video understanding?}

Despite the existence of dozens of evaluation benchmarks in video understanding, there remains a significant domain gap between these evaluations and real-world video understanding tasks. 
Early evaluations~\cite{xu2017video,yu2019activitynet,jang2017tgif} are largely based on video understanding and retrieval datasets~\cite{xu2016msr,caba2015activitynet}, assessing models through coarse-grained QA tasks, such as \textit{``Q: Who is dancing? A: Man''}. 
These QAs predominantly focus on short videos with fixed question types and lack temporal indispensability~\cite{fang2024mmbench}. Subsequent works~\cite{li2024mvbench,fu2024video,zhou2024mlvu} attempt to address these limitations by extending video temporal length and incorporating more diverse tasks and video sources. E.T.Bench~\cite{liu2024etbench} advances this further by exploring inherent temporal information in videos and evaluating fine-grained temporal event detection capabilities. 
However, all the aforementioned works are limited to offline settings, where models have access to all video frames when answering queries. While these models exhibit impressive performance on offline video understanding benchmarks, a substantial gap remains between their demonstrated capabilities and the requirements of a real-world assistant or autonomous agent.

A pioneering benchmark, VStream-QA~\cite{zhang2024flashvstreammemorybasedrealtimeunderstanding}, represents one of the earliest efforts to evaluate streaming understanding, leveraging video sources from Ego4d~\cite{grauman2022ego4dworld3000hours} and MovieNet~\cite{huang2020movienet}. Meanwhile, StreamingBench~\cite{lin2024streamingbench}, a most recent work, expands the scope by evaluating Video-LLMs on a larger scale in streaming scenarios. However, three primary evaluation categories of StreamingBench primarily target the leverage of existing visual inputs to respond to incoming queries immediately, resulting in an incomplete portrayal of streaming perception. 

In this work, we propose that effective online video understanding requires simultaneous capabilities to
\textbf{trace back past information}, \textbf{perceive the going-on}, and \textbf{forward active responding} simultaneously. 
Given a query during a streaming video, a Video-LLM must determine whether to respond immediately using past and ongoing information or wait until sufficient evidence has been accumulated. We refer to this as the \textbf{Video Chain-of-Time} thinking process (Figure~\ref{fig:pipeline}), inspired by the Chain-of-Thought reasoning in LLMs~\cite{wei2023chainofthoughtpromptingelicitsreasoning}.

We introduce \textbf{OVO-Bench} (\textbf{O}nline-\textbf{V}ide\textbf{O}-\textbf{Bench}mark) to evaluate Video-LLMs' online video understanding capabilities. The benchmark comprises 644 videos from diverse sources, including curated datasets and web videos, spanning 7 major domains (Sports, Video Games, Ego Centric, \textit{etc.}) with durations ranging from minutes to half an hour. Using a hybrid approach combining semi-automated MLLM generation and human curation, we created 2814 high-quality samples (\textbf{Meta-Annotations}) with precise event timestamps. These Meta-Annotations are organized into 12 tasks across three categories: \textbf{Backward Tracing}, \textbf{Real-Time Visual Perception}, and \textbf{Forward Active Responding}, reflecting the human video understanding process illustrated in Fig.~\ref{fig:pipeline}.
Notably, the proposed \textbf{Forward Active Responding} marks the \textbf{first} evaluation that requires models to continuously adapt their responses to ongoing visual input for online video understanding.

Building on the human-reviewed meta-annotations, we develop an evaluation pipeline that queries Video-LLMs densely along temporal axes to simulate continuous information processing. For \textbf{Backward Tracing} and \textbf{Real-Time Visual Perception}, we adopt multiple-choice evaluation, converting videos into segments from start to query time to accommodate offline models. With this approach, we explore the potential of explicitly leveraging state-of-the-art offline Video-LLMs for online video understanding. We evaluated eleven Video-LLMs, including proprietary models GPT-4o~\cite{openai2023gpt4} and Gemini-1.5-Pro~\cite{team2024gemini}, alongside six recent open-source MLLMs like Qwen2-VL~\cite{Qwen2VL} and LLaVA-OneVision~\cite{li2024llavaonevision}. 
Despite their strong offline performance, these models struggle with online-style queries  (e.g., \textit{What is happening now?}), showing a significant gap from human performance. Further experiments on recent streaming models, such as Flash-VStream~\cite{zhang2024flashvstreammemorybasedrealtimeunderstanding}, reveal an even wider performance gap compared to offline counterparts, highlighting a substantial research space for further exploration and improvement.
\begin{figure*}[ht]
    \centering
    \includegraphics[width=1\textwidth]{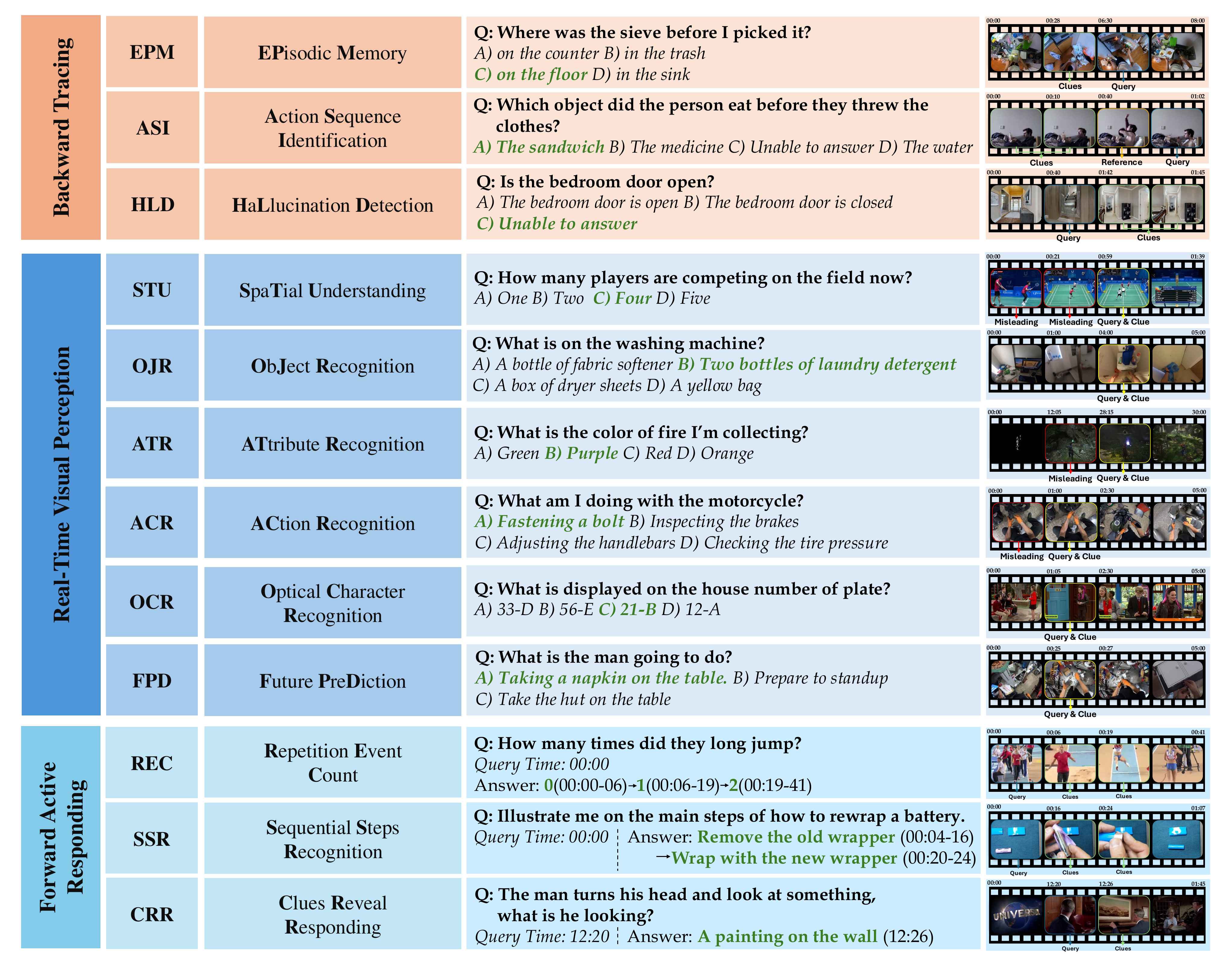}
    \vspace{-18pt}
    \caption{\textbf{Examples of each task in OVO-Bench.} The 14 tasks are categorized into three different kinds of perceiving modes in online video understanding: \textbf{Backward Tracing}, \textbf{Real-Time Visual Perception}, and \textbf{Forward Active Responding}.}
    \label{fig:pdfpart}
    \vspace{-15pt}
\end{figure*}

\section{Related Works}
\label{sec:formatting}

\noindent\textbf{Video Large Language Models.} 
\noindent Video  Large Language Models (VLLMs) can process a video by treating it as a sequence of video frames. Projects like VideoChat~\cite{li2023videochat}, Video-LLaMA~\cite{damonlpsg2023videollama}, and Video-ChatGPT~\cite{Maaz2023VideoChatGPT} project the CLIP-ViT~\cite{radford2021learningtransferablevisualmodels} embeddings of selected video frames through a Multi-Layer Perceptron (MLP) projector into the LLM embedding space, then concatenate these embeddings with text embeddings for enhanced video understanding. However, the context length of MLLMs limits their effectiveness in understanding long videos~\cite{li2023videochat,Maaz2023VideoChatGPT}, as longer videos require more frames and a longer context length. To address this limitation, two major approaches have been developed: compressing video features and selecting critical frames.

In the realm of feature compression, Chat-UniVi~\cite{jin2023chatunivi} merges similar visual tokens through clustering techniques. MovieChat~\cite{song2023moviechat} and MA-LLM~\cite{mallm} employ a memory bank to store a fixed number of video tokens by iteratively merging the most similar tokens. ST-LLM~\cite{liu2025st} and MovieChat~\cite{song2023moviechat} reduce video tokens to 32 using a pretrained Q-Former from BLIP2~\cite{dai2023instructblip}. LLaMA-VID~\cite{li2024llamavid} takes a more radical approach, compressing each frame into a content token and a context token.

On the other hand, frame selection methods aim to identify the most representative frames. VideoStreaming~\cite{qian2024streaming} utilizes a small LLM to select critical video clips, while FlashVstream~\cite{zhang2024flashvstreammemorybasedrealtimeunderstanding} employs a clustering method to choose representative frames for high-resolution processing. LongVU~\cite{shen2024longvu} leverages question embeddings to select question-related frames, thereby enhancing video understanding.

\noindent\textbf{Benchmarks for Video Understanding.} 
\noindent Traditional video benchmarks, \textit{e.g.,} MSVD-QA~\cite{xu2017video}, MSRVTT-QA~\cite{xu2017video}, and ActivityNet-QA~\cite{yu2019activitynet}, predominantly consist of short videos, typically ranging from 1 to 2 minutes in duration. These datasets are meticulously annotated with corresponding questions and ground truth answers. GPT-4~\cite{openai2023gpt4} is employed to assess the accuracy of the answers by comparing them against the provided questions and ground truth responses. However, these benchmarks primarily focus on evaluating short, static video scenes. Hence, new benchmarks designed to test causal and temporal understanding, \textit{e.g.,} NExT-QA~\cite{xiao2021next}, TemporalBench~\cite{cai2024temporalbenchbenchmarkingfinegrainedtemporal}, and AutoEval-Video~\cite{chen2024autoevalvideoautomaticbenchmarkassessing} are proposed.

To gauge the capabilities of models on long-duration videos, benchmarks like EgoSchema~\cite{mangalam2023egoschemadiagnosticbenchmarklongform} covering over 5,000 egocentric videos with an average length of 3 minutes have been introduced. In contrast, Video-MME~\cite{fu2024video}, LVBench~\cite{wang2024lvbench}, and LongVideoBench~\cite{wu2024longvideobenchbenchmarklongcontextinterleaved} feature videos spanning from 20 minutes to over an hour, evaluating a broad spectrum of video understanding capabilities. HourVideo~\cite{chandrasegaran2024hourvideo} stands out with egocentric videos extending up to 2 hours, accompanied by more than 12,976 multiple-choice questions. Unlike these offline video benchmarks, our proposed \textbf{OVO-Bench} is designed to evaluate online, interactive video understanding.

\noindent\textbf{Online Video Understanding.}
\noindent Traditional offline video understanding methodologies primarily focus on accessing entire video sequences to facilitate prediction tasks. Conversely, online video understanding demands models to process video streams sequentially, making decisions based on current and past information. This approach is particularly well-suited for scenarios where future data is unavailable, such as in embodied intelligence, autonomous driving, and augmented reality applications. Among online video understanding methods~\cite{zhang2024internlmol,qian2025dispider}, 
FlashVStream~\cite{zhang2024flashvstreammemorybasedrealtimeunderstanding} employs a clustering method to select representative frames, enabling MLLMs for real-time interactions. LIVE~\cite{videollm-online} introduces a comprehensive framework for learning in video streams, which includes a training objective, data generation schema, and an inference pipeline for online video understanding.

\section{OVO-Bench}

In this section, we present the construction process of our OVO-Bench. We start with a detailed introduction to the three different modes of online video understanding, followed by a comprehensive description of the data collection and annotation procedures. A statistical report of our proposed benchmark is displayed at the end of this section.
\subsection{Online Video Understanding Mode Taxonomy}
Online video understanding aims to equip real-world, always-on agents with the ability to receive and process video inputs continuously, which closely mimics the human visual perception process. We categorize online video understanding into three distinct problem-solving modes: \textbf{(1) Backward Tracing}, \textbf{(2) Real-Time Visual Perception}, and \textbf{(3) Forward Active Responding}. Given a user-provided text query \textbf{$Q_{t_0}$} at the current time \textbf{$t_{0}$} and a streaming video input \textbf{$X_{(-\infty, +\infty)}$}, these modes are formally defined as follows:
\begin{enumerate}
    \item \textbf{Backward Tracing}:
    \[
    R_{t_0} = P(Q_{t_0}, X_{(-\infty, -T]})
    \]
    
    \item \textbf{Real-Time Visual Perception}:
    \[
    R_{t_0} = P(Q_{t_0}, X_{(-T, t_{0}]})
    \]
    
    \item \textbf{Forward Active Responding}:
    \[
    R_{(t_0, +\infty)} = P(Q_{t_0}, X_{(t_0, +\infty)})
    \]
\end{enumerate}
in which \textbf{$T$} represents a threshold that defines the boundary for recent times, and $R$ denotes the model's response. The first two modes, \textit{Backward Tracing} and \textit{Real-Time Visual Perception}, involve collecting visual information from past and current timeframes respectively, and are expected to give immediate responses. In contrast, \textit{Forward Active Responding} requires the model to withhold a response until sufficient future information becomes available to ensure a confident answer. Based on these distinctions, we have meticulously designed tasks tailored to each mode to effectively evaluate the performance of Video-LLMs across these diverse capabilities.
\subsubsection{Backward Tracing}
Memory, particularly long-term memory, is a crucial aspect of human intelligence. In video understanding systems, this capability involves recalling and reasoning about past events. We focus on the following three tasks to evaluate this capability:
\begin{enumerate}
    \item \textbf{[EPM] Episodic Memory:} Backtrack and retrieve key moments from past video inputs. 

    \item \textbf{[ASI] Action Sequence Identification:} Identify the correct sequence of human actions in the video streams.

    \item \textbf{[HLD] Hallucination Detection:} Ask questions irrelevant to existing video inputs.
\end{enumerate}


\subsubsection{Real-Time Visual Perception}
Accurate real-time perception of visual content is crucial, as actions undertaken in the present shape future outcomes. In various real-world scenarios, an immediate and precise understanding of ongoing visual inputs is essential. We propose six critical categories that constitute the foundational capabilities for effective real-time visual perception:
\begin{enumerate}
    \item \textbf{[STU] Spatial Understanding.} Reason over the spatial relationships between objects occurring in nearby frames.

    \item \textbf{[OJR] Object Recognition.} Recognize the objects appearing in the current frames.

    \item \textbf{[ATR] Attribute Recognition.} Identify the characteristics or properties of objects, such as color, texture, and size that appear in nearby frames.

    \item \textbf{[ACR] Action Recognition.} Recognize and interpret the actions being performed by individuals in the current frame.

    \item \textbf{[OCR] Optical Character Recognition.} Recognize and interpret characters that appear within the frame.

    \item \textbf{[FPD] Future Prediction.} Forecast the most probable subsequent phase of the current scene, including changes in object states, actions, and other dynamic elements.
\end{enumerate}

\subsubsection{Forward Active Responding}
Transitioning from passive reception to active perception is essential for advanced video understanding systems. Existing benchmarks primarily focus on the aforementioned two understanding modes, where Video-LLMs are required to respond immediately based on available information. In contrast, we introduce the \emph{Forward Active Responding} mode, which allows the model to adjust its responses based on forthcoming visual inputs. We devise three task dimensions to evaluate the models' active responding abilities:
\begin{enumerate}
    \item \textbf{[REC] Repetition Event Count.} Respond when a repetitive event occurs again, including both high-frequency repetitive actions over short durations and semantically long-term repetitive occurrences of certain events.

    \item \textbf{[SSR] Sequential Steps Recognition.} Respond when a certain procedure or sequence of actions has transitioned to another stage.

    \item \textbf{[CRR] Clues Reveal Responding.} Delay responding until sufficient information or clues are provided.
\end{enumerate}

\begin{figure*}[ht]
    \centering
    \vspace{-20pt}
    \includegraphics[width=0.68\textwidth]{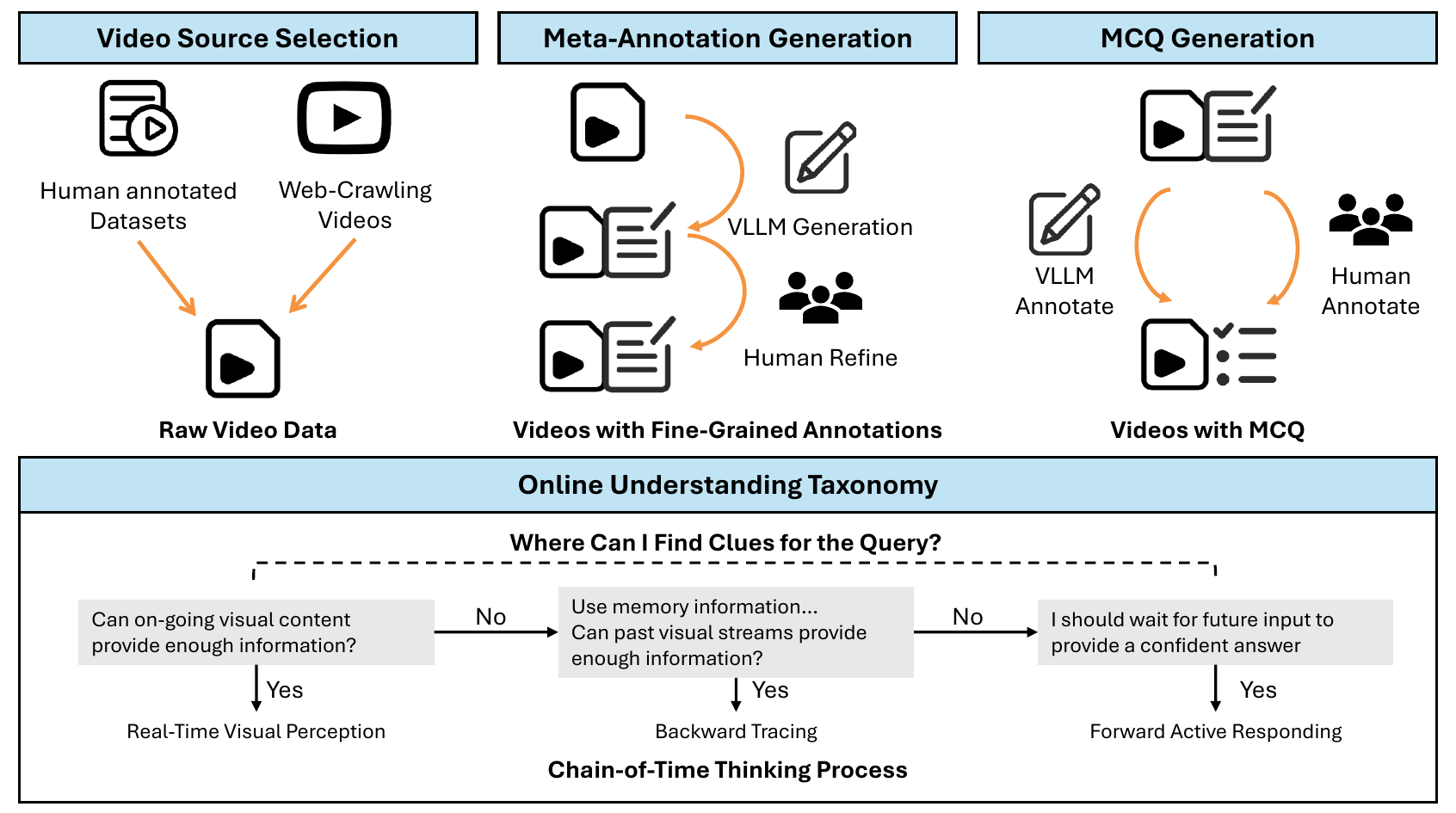}
    \vspace{-0.3cm}
    \caption{
    \textbf{Generation pipeline of OVO-Bench.} 
    Within public annotations, 
    data is carefully filtered and relevant multiple-choice QAs are auto-generated.
    The effective system prompt and efficient answer prompt are employed to guide MLLMs toward precise outputs. The Video-LLMs we use to annotate videos are GPT-4o and Gemini-1.5 Pro.}
    \label{fig:pipeline}
\end{figure*}

\subsection{Benchmark Construction}
Under the taxonomy guidelines above, we make our first step by collecting video data and annotations from existing datasets and crawling data from the web to increase diversity. As our proposed evaluation pipeline highly relies on the accurate timestamp annotations of the referred events in the constructed prompt, the scarcity of event-level timestamps in existing datasets \cite{wu2024star}\cite{majumdar2024openeqa}\cite{patraucean2024perception} promotes the design of our highly efficient meta-data generation pipeline \ref{fig:pipeline}. Raw annotations with coarse timestamps are then refined by humans to ensure accuracy. Our final questions and options for evaluation are constructed using our rule-based pipeline based on these human-refined meta-annotations. All QA samples undergo manual inspection before being included in the final test set.
\subsubsection{Video and Annotation Collection}
\textbf{Video Source Selection.} We follow existing benchmarks \cite{liu2024etbench}\cite{li2024mvbench} by exploiting high-quality customized video datasets, and enrich our diversity by utilizing self-crawling videos from different domains. \textbf{(1) Human-annotated Video Dataset.} Our main consideration for utilizing organized datasets is to alleviate the labor-intensive source video collection process. Specifically, we include QA-Ego4D\cite{barmann2022did} and OpenEQA\cite{majumdar2023openeqa} for the \textbf{[EPM]} task, STAR\cite{wu2024star}, YouCook2\cite{youcook2}, CrossTask\cite{zhukov2019cross}, HiREST\cite{zala2023hierarchical}, and COIN\cite{tang2019coin} for the \textbf{[ASI]} task, Perception-Test\cite{patraucean2024perception} and Thumos\cite{jiang2014thumos}\cite{gorban2015thumos} for the \textbf{[REC]} task, COIN\cite{tang2019coin} for the \textbf{[SSR]} task, MovieNet\cite{huang2020movienet} for the \textbf{[CRR]} task, and Ego4D\cite{grauman2022ego4dworld3000hours} for tasks under \textit{Real-Time Visual Perception}. All samples are selected from val or test sets to avoid potential data leakage. \textbf{(2) Web-crawling Videos.} To further extend the diversity of our benchmark, we follow the existing practice \cite{fang2024mmbench}\cite{lin2024streamingbench} of crawling source videos from YouTube.
\\
\textbf{Meta-Annotations Collection.} We employ three approaches to collect our meta-annotations, which contain event-level timestamps: \textbf{(1) Existing Annotation Repurposing.} For human-annotated datasets with accurate event-level timestamps \cite{barmann2022did}\cite{tang2019coin}\cite{grauman2022ego4dworld3000hours}, we explicitly take advantage of these labels and reconstruct them to our final prompt. \textbf{(2) Semi-Automatic Generation.} For datasets that provide video-level QA pairs without complete temporal localization, including \cite{majumdar2024openeqa}\cite{wu2024star}\cite{patraucean2024perception}\cite{jiang2014thumos}\cite{gorban2015thumos}, we prompt temporal-sensitive Video-LLMs like Gemini-1.5\cite{team2024gemini} to provide coarse-grain timestamps which fit the event referred in question and answer. For tasks under the \textit{Real-Time Visual Perception} scenario, timestamps are given during our automatic QA construction process, which will be illustrated in \ref{sec:QA_Generation}. \textbf{(3) Human-annotated.} For the \textbf{[SSR]} and \textbf{[CRR]} tasks, questions, answers, and ground-truth timestamps are collected by our recruited volunteers. We then perform a meticulous inspection of all collected source videos and the corresponding meta-annotations to ensure precision. 
\\

\begin{figure*}[ht]
    \centering
    \includegraphics[width=0.32\textwidth]{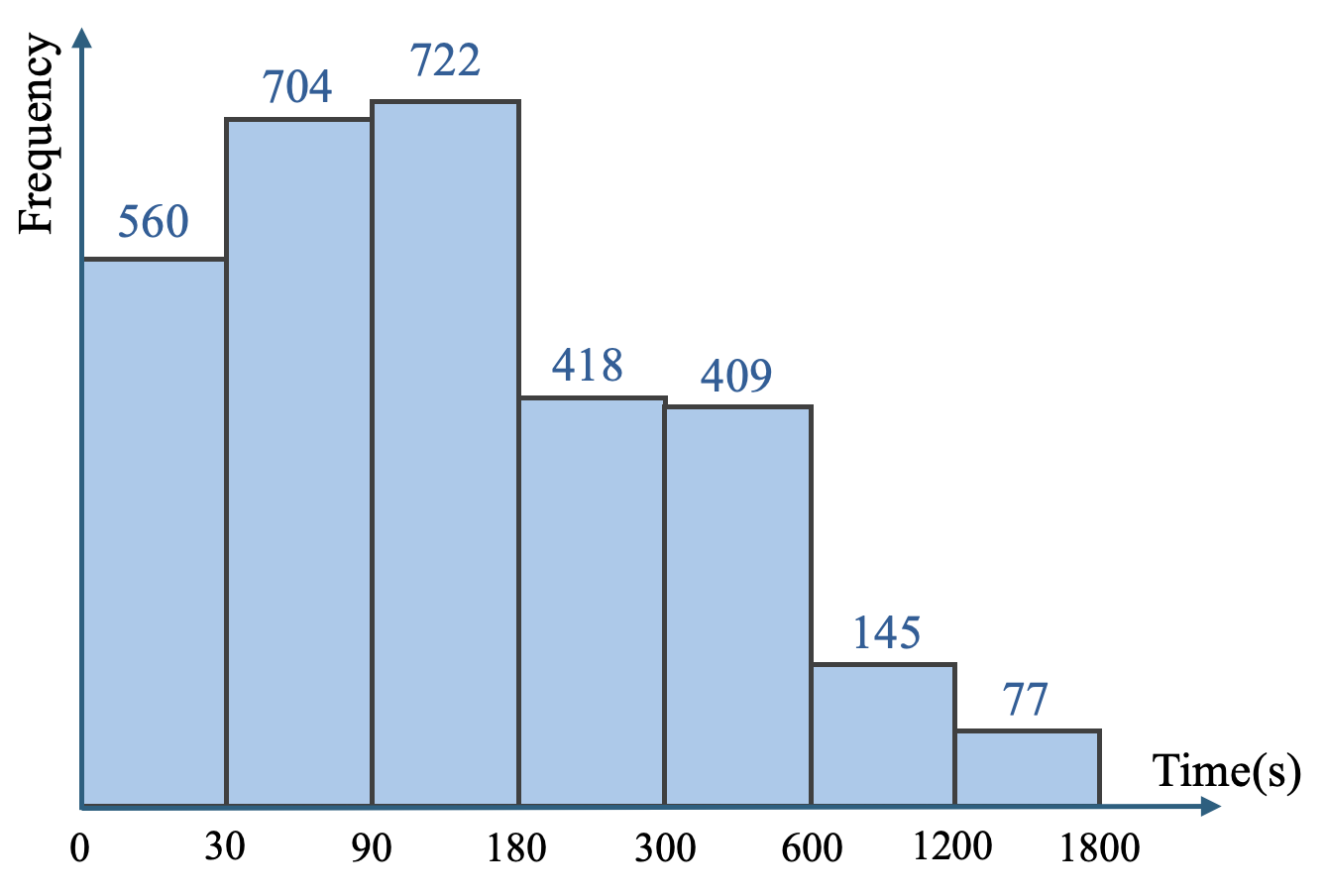}
    \hfill 
    \includegraphics[width=0.42\textwidth]{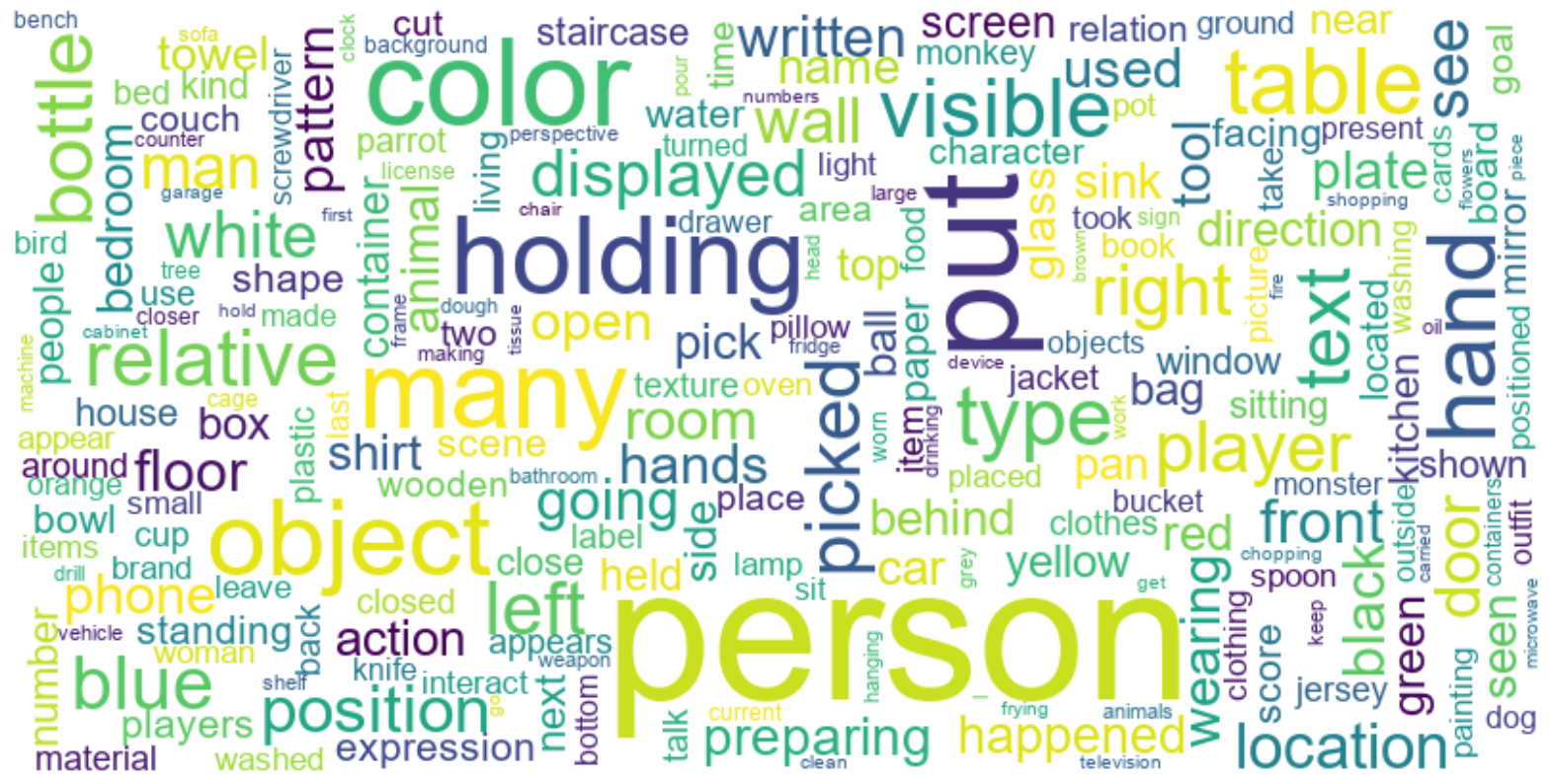}
    \hfill 
    \includegraphics[width=0.24\textwidth]{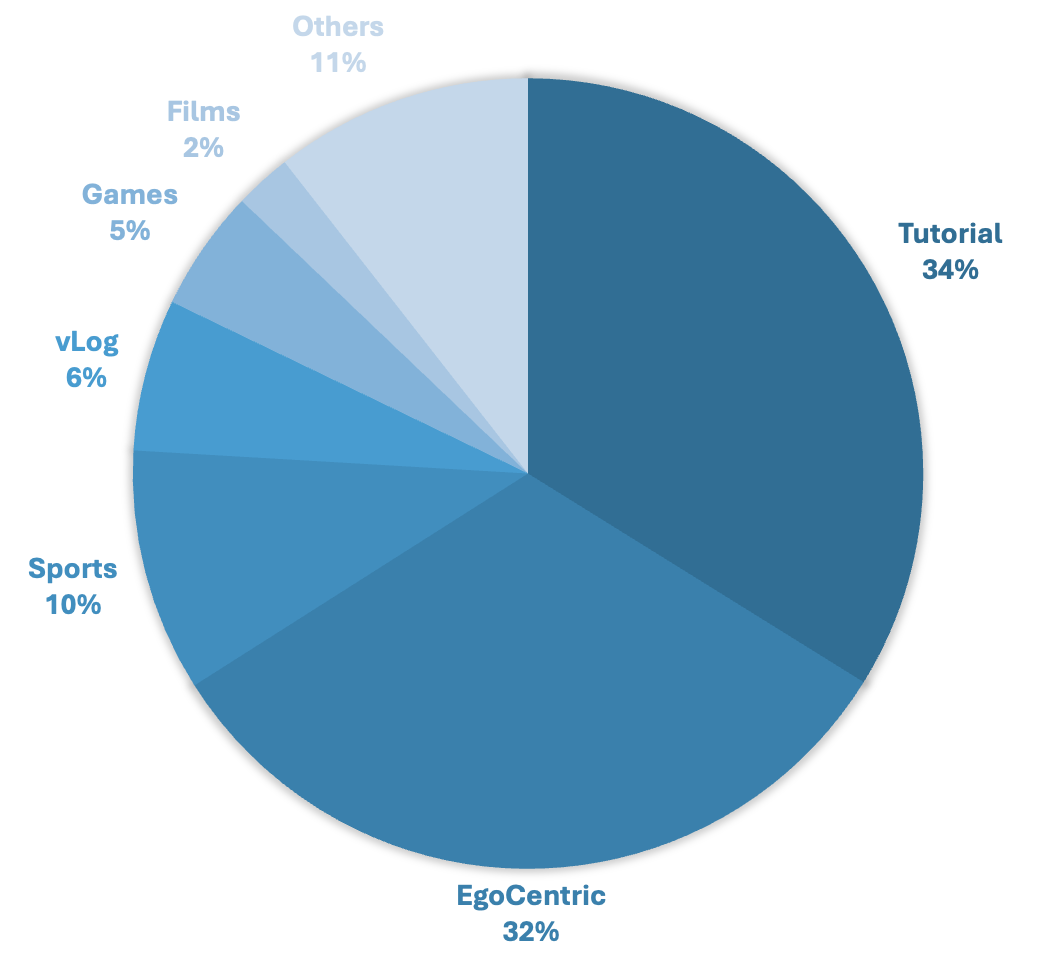} 
    \vspace{-0.3cm}
    \caption{
    \textbf{Left}: Queries Temporal Distribution in OVO-Bench. \textbf{Center}: Linguistic Characteristics of Text Queries. \textbf{Right}: Video category distribution of OVO-Bench.}
    \label{fig:data}
\end{figure*}

\subsubsection{Prompt Generation}\label{sec:QA_Generation}
\textbf{Question and Answer Generation.} Besides carefully selecting QA pairs from existing datasets to fit into our proposed tasks, we also adopt a highly efficient automatic question and answer generation pipeline, particularly for the \textit{Real-Time Visual Perception} scenario. We randomly sample short clips from original long-form videos and then leverage GPT-4o\cite{hurst2024gpt} to select potential candidates and construct questions and corresponding answers using human-refined prompts. Human-proposed questions are also adopted as a part of these tasks to alleviate possible LLM preferences. For the novel \textbf{[CRR]} task, even the strongest Video-LLMs/MLLMs like Gemini-1.5-Pro struggle to construct desired problems. Volunteers are then recruited to provide QA pairs under our guidance.
\\
\textbf{Options Generation and Selection.} 
We adopt multiple-choice questions as testing forms for \textit{Backward Tracing} and \textit{Real-Time Visual Perception} scenarios. However, as revealed in \cite{chen2024we}, the naively designed options of a multi-choice form query can cause information leakage about answers. We propose to generate options using a carefully designed rule-based and visually grounded transformation of correct answers, bringing misleading information from original videos to increase difficulty. Specifically, we prompt Video-LLMs with original QA pairs and corresponding video clips to generate visual-related options. A careful human review is then conducted to further ensure the options' effectiveness. All options are shuffled after human review to avoid potential preference bias.
\\
\textbf{Prompting Offline-Models for Simulated Online Understanding.}
With the significant performance gap between main-streaming powerful offline Video-LLMs \cite{team2024gemini}\cite{openai2024gpt4o}\cite{Qwen2VL} and existing online models \cite{videollm-online}\cite{zhang2024flashvstreammemorybasedrealtimeunderstanding}, one natural question is made: \textit{Is it effective to prompt offline models directly for online video understanding?} For the \textit{Real-Time Visual Perception} setting, we make human curation to the original question to include implies about the real-time query scenarios, for example, by using sentence patterns like \textit{What is/What am I} or containing words like \textit{Now/Currently}. We made another intuitive attempt to prompt offline models to solve tasks under our novel \textit{Forward Active Responding} scenario, which asks for a continuous adapting capability. Specifically, we devise a multiple-triggering densely query and evaluation pipeline, allowing the model to decide whether existing information has provided enough clues to answer the user's query. 

\subsection{Datasets Statistics}

OVO-Bench consists of 644 unique videos spanning 7 major domains, including Sports, Video Games, and Tutorial, among others. The video durations range from a few minutes to half an hour, with the average query timepoint being 428.89 seconds. Figure~\ref{fig:data} \textbf{Left} illustrates the duration distribution of the queries within OVO-Bench. The benchmark includes 2,814 question-answer (QA) pairs, featuring a large number of multiple-choice questions and a smaller set of open-ended questions. The number of options for the multiple-choice questions varies between 2 and 5, rather than being fixed at four. The distribution of video category is visualized in Figure~\ref{fig:data}~\textbf{Right}.

\begin{table*}[h]
    \centering
    \renewcommand{\arraystretch}{1.2}  
    \resizebox{1.0\textwidth}{!}{%
\begin{tabular}{lccccccccccccccccc}
\hline\hline
\multicolumn{1}{c|}{} &
  \multicolumn{1}{l|}{} &
  \multicolumn{7}{c|}{\textbf{Real-Time Visual Perception}} &
  \multicolumn{4}{c|}{\textbf{Backward Tracing}} &
  \multicolumn{4}{c|}{\textbf{Forward Active Responding}} &
  \textbf{Overall Avg.} \\ \cline{3-18} 
\multicolumn{1}{c|}{\multirow{-2}{*}{\textbf{Model}}} &
  \multicolumn{1}{l|}{\multirow{-2}{*}{\textbf{\# Frames}}} &
  OCR &
  ACR &
  ATR &
  STU &
  FPD &
  \multicolumn{1}{c|}{OJR} &
  \multicolumn{1}{c|}{Avg.} &
  EPM &
  ASI &
  \multicolumn{1}{c|}{HLD} &
  \multicolumn{1}{c|}{Avg.} &
  REC &
  SSR &
  \multicolumn{1}{c|}{CRR} &
  \multicolumn{1}{c|}{Avg.} &
  Overall Avg. \\ \hline
\multicolumn{18}{c}{\textbf{Human}} \\ \hline
\multicolumn{1}{l|}{Human Agents} &
  \multicolumn{1}{c|}{-} &
  93.96 &
  92.57 &
  94.83 &
  92.70 &
  91.09 &
  \multicolumn{1}{c|}{94.02} &
  \multicolumn{1}{c|}{93.20} &
  92.59 &
  93.02 &
  \multicolumn{1}{c|}{91.37} &
  \multicolumn{1}{c|}{92.33} &
  95.48 &
  89.67 &
  \multicolumn{1}{c|}{93.56} &
  \multicolumn{1}{c|}{92.90} &
  92.81 \\ \hline
\multicolumn{18}{c}{\textbf{Blind LLMs}} \\ \hline
\multicolumn{1}{l|}{GPT-4-turbo\cite{openai2023gpt4}} &
  \multicolumn{1}{c|}{-} &
  28.86 &
  24.77 &
  25.67 &
  33.76 &
  27.72 &
  \multicolumn{1}{c|}{26.63} &
  \multicolumn{1}{c|}{27.90} &
  42.76 &
  48.65 &
  \multicolumn{1}{c|}{70.05} &
  \multicolumn{1}{c|}{53.82} &
  - &
  - &
  \multicolumn{1}{c|}{52.92} &
  \multicolumn{1}{c|}{-} &
  - \\ \hline
\multicolumn{18}{c}{\textbf{Proprietary Multimodal Models-Offline}} \\ \hline
\multicolumn{1}{l|}{Gemini 1.5 Pro\cite{team2024gemini}} &
  \multicolumn{1}{c|}{1fps} &
  \textbf{85.91} &
  \textbf{66.97} &
  \textbf{79.31} &
  \textbf{58.43} &
  63.37 &
  \multicolumn{1}{c|}{\textbf{61.96}} &
  \multicolumn{1}{c|}{\textbf{69.32}} &
  \textbf{58.59} &
  \textbf{76.35} &
  \multicolumn{1}{c|}{\textbf{52.64}} &
  \multicolumn{1}{c|}{\textbf{62.54}} &
  \textbf{35.53} &
  \textbf{74.24} &
  \multicolumn{1}{c|}{\textbf{61.67}} &
  \multicolumn{1}{c|}{\textbf{57.15}} &
  \textbf{63.00} \\
\multicolumn{1}{l|}{GPT-4o\cite{openai2024gpt4o}} &
  \multicolumn{1}{c|}{64} &
  69.8 &
  64.22 &
  71.55 &
  51.12 &
  \textbf{70.3} &
  \multicolumn{1}{c|}{59.78} &
  \multicolumn{1}{c|}{64.46} &
  57.91 &
  75.68 &
  \multicolumn{1}{c|}{48.66} &
  \multicolumn{1}{c|}{60.75} &
  27.58 &
  73.21 &
  \multicolumn{1}{c|}{59.4} &
  \multicolumn{1}{c|}{53.40} &
  59.54 \\ \hline
\multicolumn{18}{c}{\textbf{Open-source Multimodal Models-Offline}} \\ \hline
\multicolumn{1}{l|}{Qwen2-VL-72B\cite{Qwen2VL}} &
  \multicolumn{1}{c|}{64} &
  65.77 &
  \textbf{60.55} &
  69.83 &
  51.69 &
  69.31 &
  \multicolumn{1}{c|}{54.35} &
  \multicolumn{1}{c|}{61.92} &
  52.53 &
  \textbf{60.81} &
  \multicolumn{1}{c|}{\textbf{57.53}} &
  \multicolumn{1}{c|}{\textbf{56.95}} &
  \textbf{38.83} &
  64.07 &
  \multicolumn{1}{c|}{45.00} &
  \multicolumn{1}{c|}{49.30} &
  \textbf{56.27} \\
\multicolumn{1}{l|}{LLaVA-Video-7B\cite{zhang2024videoinstructiontuningsynthetic}} &
  \multicolumn{1}{c|}{64} &
  \textbf{69.13} &
  58.72 &
  68.83 &
  49.44 &
  \textbf{74.26} &
  \multicolumn{1}{c|}{59.78} &
  \multicolumn{1}{c|}{63.52} &
  \textbf{56.23} &
  57.43 &
  \multicolumn{1}{c|}{7.53} &
  \multicolumn{1}{c|}{40.4} &
  34.10 &
  \textbf{69.95} &
  \multicolumn{1}{c|}{60.42} &
  \multicolumn{1}{c|}{\textbf{54.82}} &
  52.91 \\
\multicolumn{1}{l|}{LLaVA-OneVision-7B\cite{li2024llavaonevision}} &
  \multicolumn{1}{c|}{64} &
  66.44 &
  57.80 &
  \textbf{73.28} &
  \textbf{53.37} &
  71.29 &
  \multicolumn{1}{c|}{\textbf{61.96}} &
  \multicolumn{1}{c|}{\textbf{64.02}} &
  54.21 &
  55.41 &
  \multicolumn{1}{c|}{21.51} &
  \multicolumn{1}{c|}{43.71} &
  25.64 &
  67.09 &
  \multicolumn{1}{c|}{58.75} &
  \multicolumn{1}{c|}{50.50} &
  52.74 \\
\multicolumn{1}{l|}{Qwen2-VL-7B\cite{Qwen2VL}} &
  \multicolumn{1}{c|}{64} &
  60.40 &
  50.46 &
  56.03 &
  47.19 &
  66.34 &
  \multicolumn{1}{c|}{55.43} &
  \multicolumn{1}{c|}{55.98} &
  47.81 &
  35.48 &
  \multicolumn{1}{c|}{56.08} &
  \multicolumn{1}{c|}{46.46} &
  31.66 &
  65.82 &
  \multicolumn{1}{c|}{48.75} &
  \multicolumn{1}{c|}{48.74} &
  50.39 \\
\multicolumn{1}{l|}{InternVL-V2-8B\cite{chen2023internvl}} &
  \multicolumn{1}{c|}{64} &
  67.11 &
  \textbf{60.55} &
  63.79 &
  46.07 &
  68.32 &
  \multicolumn{1}{c|}{56.52} &
  \multicolumn{1}{c|}{60.39} &
  48.15 &
  57.43 &
  \multicolumn{1}{c|}{24.73} &
  \multicolumn{1}{c|}{43.44} &
  26.5 &
  59.14 &
  \multicolumn{1}{c|}{54.14} &
  \multicolumn{1}{c|}{46.60} &
  50.15 \\
\multicolumn{1}{l|}{LongVU-7B\cite{shen2024longvu}} &
  \multicolumn{1}{c|}{1fps} &
  53.69 &
  53.21 &
  62.93 &
  47.75 &
  68.32 &
  \multicolumn{1}{c|}{59.78} &
  \multicolumn{1}{c|}{57.61} &
  40.74 &
  59.46 &
  \multicolumn{1}{c|}{4.84} &
  \multicolumn{1}{c|}{35.01} &
  12.18 &
  69.48 &
  \multicolumn{1}{c|}{\textbf{60.83}} &
  \multicolumn{1}{c|}{47.50} &
  46.71 \\ \hline
\multicolumn{18}{c}{\textbf{Open-source Multimodal Models-Online}} \\ \hline
\multicolumn{1}{l|}{Flash-VStream-7B\cite{zhang2024flashvstreammemorybasedrealtimeunderstanding}} &
  \multicolumn{1}{c|}{1fps} &
  24.16 &
  29.36 &
  28.45 &
  33.71 &
  25.74 &
  \multicolumn{1}{c|}{28.80} &
  \multicolumn{1}{c|}{28.37} &
  39.06 &
  37.16 &
  \multicolumn{1}{c|}{5.91} &
  \multicolumn{1}{c|}{27.38} &
  8.02 &
  \textbf{67.25} &
  \multicolumn{1}{c|}{\textbf{60.00}} &
  \multicolumn{1}{c|}{\textbf{45.09}} &
  33.61 \\
\multicolumn{1}{l|}{VideoLLM-online-8B\cite{videollm-online}} &
  \multicolumn{1}{c|}{2fps} &
  8.05 &
  23.85 &
  12.07 &
  14.04 &
  45.54 &
  \multicolumn{1}{c|}{21.20} &
  \multicolumn{1}{c|}{20.79} &
  22.22 &
  18.80 &
  \multicolumn{1}{c|}{\textbf{12.18}} &
  \multicolumn{1}{c|}{17.73} &
  - &
  - &
  \multicolumn{1}{c|}{-} &
  \multicolumn{1}{c|}{-} &
  - \\ 
\multicolumn{1}{l|}{Dispider\cite{qian2025dispider}} &
  \multicolumn{1}{c|}{1fps} &
  \textbf{57.72} &
  \textbf{49.54} &
  \textbf{62.07} &
  \textbf{44.94} &
  \textbf{61.39} &
  \multicolumn{1}{c|}{\textbf{51.63}} &
  \multicolumn{1}{c|}{\textbf{54.55}} &
  \textbf{48.48} &
  \textbf{55.41} &
  \multicolumn{1}{c|}{4.3} &
  \multicolumn{1}{c|}{\textbf{36.06}} &
  \textbf{18.05} &
  37.36 &
  \multicolumn{1}{c|}{48.75} &
  \multicolumn{1}{c|}{34.72} &
  \textbf{41.78} \\ 
\hline\hline
\end{tabular}
}\vspace{-8pt}
    \caption{\textbf{Detailed evaluation results on OVO-Bench.} To enhance the challenge of the questions by increasing the time interval between the question and the clues, the question time for \textbf{[EPM]} and \textbf{[ASI]} in the table is uniformly placed at the end of the video. For \textbf{Forward Active Responding}, accuracy-based evaluation metrics are utilized in this table. }
    \label{tab:mainresult}
\vspace{-17pt}
\end{table*}

\section{Experiments}
This section presents comprehensive experiments and in-depth analyses of OVO-Bench. 

\subsection{Models and Evaluation Strategies}
We evaluate four existing types of models: (1) Offline Multimodal Models, including GPT-4o~\cite{openai2024gpt4o}, Gemini-1.5-Pro~\cite{team2024gemini}, Qwen2-VL~\cite{Qwen2VL}, LLaVA-Video~\cite{zhang2024videoinstructiontuningsynthetic}, LLaVA-OneVision~\cite{liu2023llava}, InternVL-V2~\cite{chen2023internvl} and LongVU~\cite{shen2024longvu}, 
(2) Online Multimodal Models, including FlashVStream~\cite{zhang2024flashvstreammemorybasedrealtimeunderstanding} , Videollm-Online\cite{videollm-online} and Dispider\cite{qian2025dispider}, 
(3) Blind LLMs, including GPT-4-turbo~\cite{openai2023gpt4}.
(4) Human Agents.
To ensure a fair comparison of model performance, we adhere to the principle of consistency by maintaining the same number of frames or frames per second (fps) across all models. 

Considering the input video length limitations for offline Video-LLMs, we adopt specialized video input methods tailored to such models. Specifically, we segment the video into clips based on the timestamps of the questions.For instance, for a question \( Q_i \) posed at timestamp \( t_i \), we extract the video clip \( \text{Video}[0:t_i] \) as the visual input. This approach simulates a streaming question-answering scenario in online video understanding.

We also conduct a runtimes study of five models, including QWen2-VL-7B \cite{Qwen2VL}, LLaVA-Video \cite{zhang2024videoinstructiontuningsynthetic}, LLaVA-OneVision \cite{li2024llavaonevision}, InternVL-V2\cite{chen2023internvl}, and FlashVStream \cite{zhang2024flashvstreammemorybasedrealtimeunderstanding}. In this setting, we randomly select 100 samples from tasks in Backward Tracing and Real-Time Visual Perception and then plot the change of average inference delay on these videos with the number of sampled frames.

\subsection{Main Results}
Table~\ref{tab:mainresult} reports the performance of eleven models under different settings on OVO-Bench, including the \textit{Real-Time Visual Perception}, \textit{Backward Tracing}, and \textit{Forward Active Responding}. 
Our evaluation brings several important findings, as follows:

    \textbf{Offline Video-LLMs' video understanding capabilities can be effectively transferred to real-time video understanding.} The results demonstrate that offline Video-LLMs, despite being designed for offline processing, perform competitively in \textit{Real-Time Visual Perception} tasks. This suggests that the advanced video comprehension abilities developed in offline settings are transferable and can enhance performance in certain online scenarios, thereby partially bridging the gap between offline and online video understanding.

    \textbf{Current Video-LLMs lack temporal prioritization when handling VQA tasks.} Existing Video-LLMs do not prioritize real-time temporal information when answering questions, leading to an inability to accurately locate the correct scene when multiple misleading scenes matching the question appear in the video stream, as shown in Fig\ref{fig:pdfpart}. Even the best current proprietary models achieve only 58.43\% and 66.97\% on \textbf{[STU]} and \textbf{[ACR]} tasks, respectively, which represents a significant gap compared to Human Agents. 
    
    
    \textbf{Hallucinations are prevalent in Video-LLMs.} The \textbf{[HLD]}  in Table~\ref{tab:mainresult} measures hallucinations in Video-LLMs~\cite{eventhallusion}, indicating that hallucinations are a significant issue, particularly in open-source and online models. Proprietary models like Gemini 1.5 Pro perform better in managing hallucinations, yet there remains a notable gap compared to human performance(52.69\% vs. 91.37\%). This problem arises due to the models' inability to fully comprehend complex visual and temporal contexts, leading to errors in interpretation and response. Addressing hallucinations is crucial for improving the reliability and accuracy of Video-LLMs in real-world applications.

\begin{figure}[h]
    \centering
    \includegraphics[width=0.85\columnwidth]{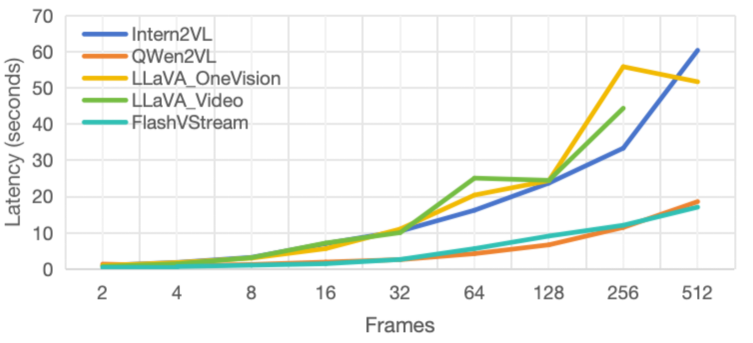}
    \vspace{-10pt}
    \caption{
    \textbf{Inference Latency (y-axis) v.s. Frames Number (x-axis).} Latency test on an A100 GPU for FlashVStream and four A100 GPUs for other models.
    }
    \label{fig:running_times}
    \vspace{-15pt}
\end{figure}

    \textbf{Current Video-LLMs need more efficient inference frameworks to achieve real-time visual question answering.} As shown in Fig\ref{fig:running_times}, the inference latencies of current Video-LLMs exhibit an exponential growth trend as frame numbers increase. Specifically, when using 64 frames as visual input, most efficient Video-LLMs, like QWen2VL-7B\cite{Qwen2VL} and FlashVStream \cite{zhang2024flashvstreammemorybasedrealtimeunderstanding}, still need around 4 seconds on average to perform a response, making real-time video dialogue far from reach.

\subsection{Comparison between online Video-LLMs and offline Video-LLMs}
\begin{figure}[h]
    \centering
    \includegraphics[width=0.65\columnwidth]{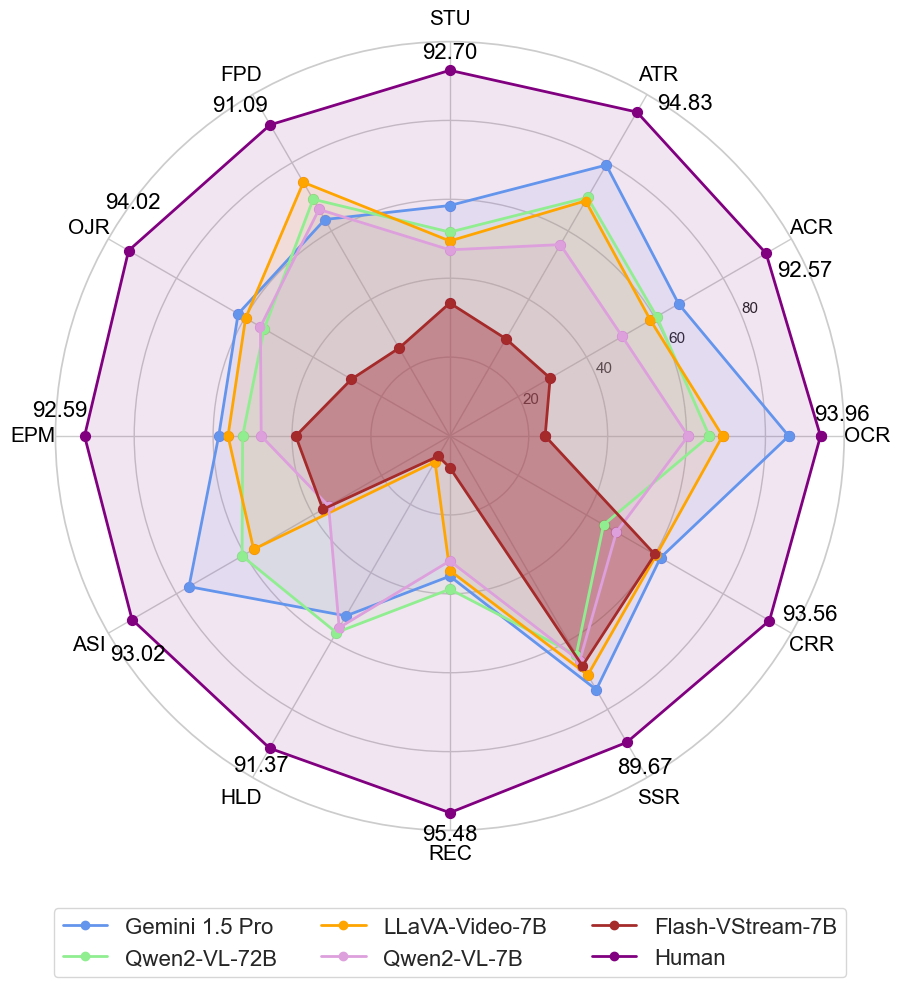}
    \vspace{-10pt}
    \caption{
    \textbf{Performance comparison between online Video-LLMs and offline Video-LLMs.} The figure illustrates the average scores of different models on the OVO-Bench in real-time visual perception tasks. }
    \label{fig:comparison}
    \vspace{-10pt}
\end{figure}

Models like \textbf{Gemini 1.5 Pro} and \textbf{Qwen2-VL-72B}, representative of offline Video-LLMs, demonstrate strong performance across various tasks, as shown in Fig\ref{fig:comparison}. Specifically, Gemini 1.5 Pro achieves the highest average score among these models. This superior performance suggests that offline models, despite not being designed for online or real-time processing, can effectively comprehend and process complex visual information when provided with sufficient computational resources and pre-processing time. Their architectures typically allow for processing the entire video sequence holistically, leveraging global context and detailed temporal information, which enhances their temporal understanding and reasoning capabilities.

In contrast, \textbf{Flash-VStream-7B}, representing online Video-LLMs, shows comparatively lower performance in real-time perception tasks compared to offline models. This model is designed to process video in a streaming manner, handling inputs frame by frame with strict latency constraints to achieve real-time responsiveness. The performance gap highlights a potential trade-off between real-time processing capabilities and the depth of visual understanding.

\subsection{Forward Active Responding}
\label{sec:far}
We include our evaluation pipeline design for our proposed \textit{Forward Active Responding}. While our high-quality human-annotated queries and clues lay an ideal testbed for future real-world online understanding models, existing naively designed online video models usually collapse in our evaluation process. We made our initial attempts to leverage our multiple-triggering query pipeline to prompt offline VideoLLMs to perform online video understanding thinking schema and further explore their potential in always-on visual perception.
\par
\begin{figure}[h]
    \centering
    \includegraphics[width=0.85\columnwidth]{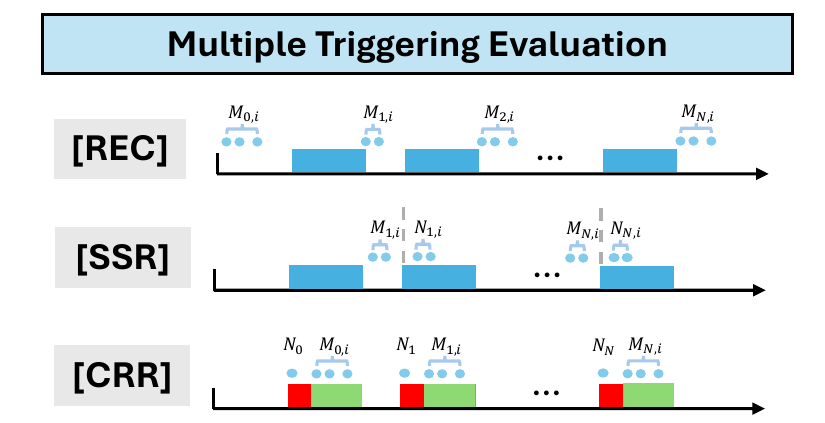}
    \vspace{-10pt}
    \caption{
    \textbf{Multiple triggering evaluation pipeline of prompt offline models for online video understanding.} Offline Video-LLMs are densely queried along the temporal axes to make independent decisions of whether existing visual content provide enough clues for answering. }
    \label{fig:evaluation_pipeline}
    \vspace{-17pt}
\end{figure}

\textbf{Evaluation Pipeline and Metrics.} As illustrated in Fig.\ref{fig:evaluation_pipeline}, We propose to query the Video-LLMs densely along the temporal axes, particularly around the interested events. Our main concerns are twofold: 1) Encourage models' timely finding of the right clues, and 2) Avoid any possible hallucination before the right clue appears. For the [REC] task, larger counting numbers are awarded. Based on this, we proposed our designed scoring metrics for the three tasks in the \textit{Forward Active Responding}.

\par
\textbf{Offline Models for Online Video Understanding.} Despite their promising performance on the \textit{Backward-Tracing} and \textit{Real-Time Visual Perception}, in which the models are given full information for making confident responses, our preliminary results show that even state-of-the-art offline models like Gemini-1.5-Pro, fails to capture the linguistic information of ongoing querying, showing limited understanding of online video content. 

\section{Conclusion and Future Work}

In this work, we introduced OVO-Bench, a comprehensive benchmark designed to assess online video understanding capabilities of Video-LLMs across three critical modes: \textit{Backward Tracing}, \textit{Real-Time Visual Perception}, and \textit{Forward Active Responding}.We anticipate that OVO-Bench will serve as a valuable resource for the research community, guiding the development of Video-LLMs toward practical, real-world applications. By highlighting current limitations and providing a platform for rigorous evaluation, we hope to inspire future research dedicated to advancing online video understanding and achieving human-level comprehension in artificial intelligence systems.

\section*{Acknowledgement}

This project is funded in part by Shanghai Artificial lntelligence Laboratory, Shanghai Innovation Institute, the National Key R\&D Program of China (2022ZD0160201). This project is also supported by the Shanghai Postdoctoral Excellence Program (No.2023023), China Postdoctoral Science Fund (No.2024M751559).
\clearpage
\setcounter{page}{1}
\maketitlesupplementary

\begin{figure*}[ht]
    \centering
    \includegraphics[width=1\textwidth]{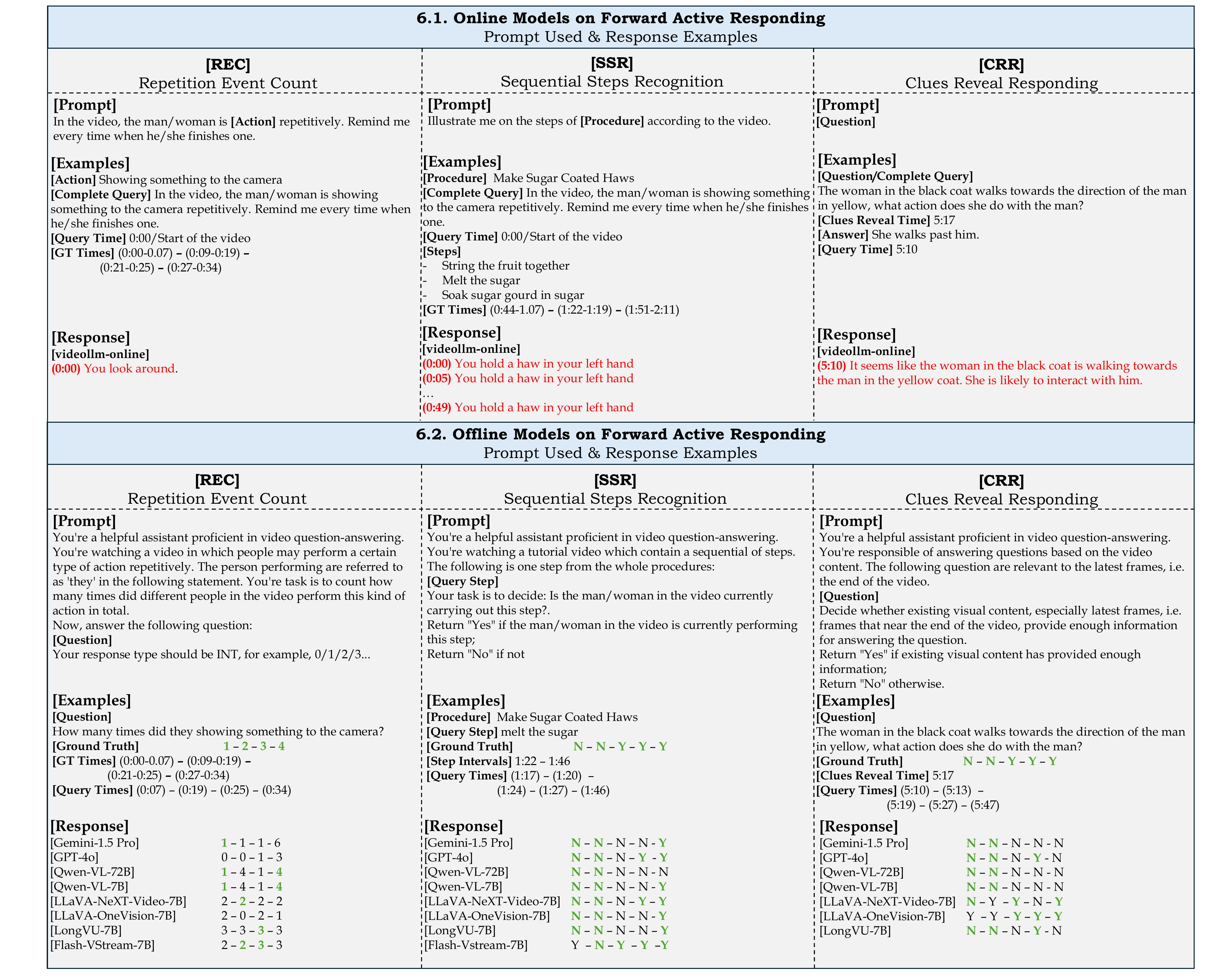}
    \vspace{-0.5cm}
    \caption{
    \textbf{Prompts used for Online(up) and Offline(down) Models on \textit{Forward Active Responding} and Response Examples.} Despite our vision for online models, existing online models, like videollm-online, are still far from satisfactory, showing limited adaptation ability, and would easily encounter collapse when processing complicated or out-of-training-domain video and queries. Offline models are inclined to perform random guessing when the queries contain words like "is/currently/ongoing".
    }
    \label{fig:case_study_forward}
    \vspace{-0.3cm}
\end{figure*}

\begin{figure*}[ht]
    \centering
    \vspace{-0.3cm}
    \includegraphics[width=1\textwidth]{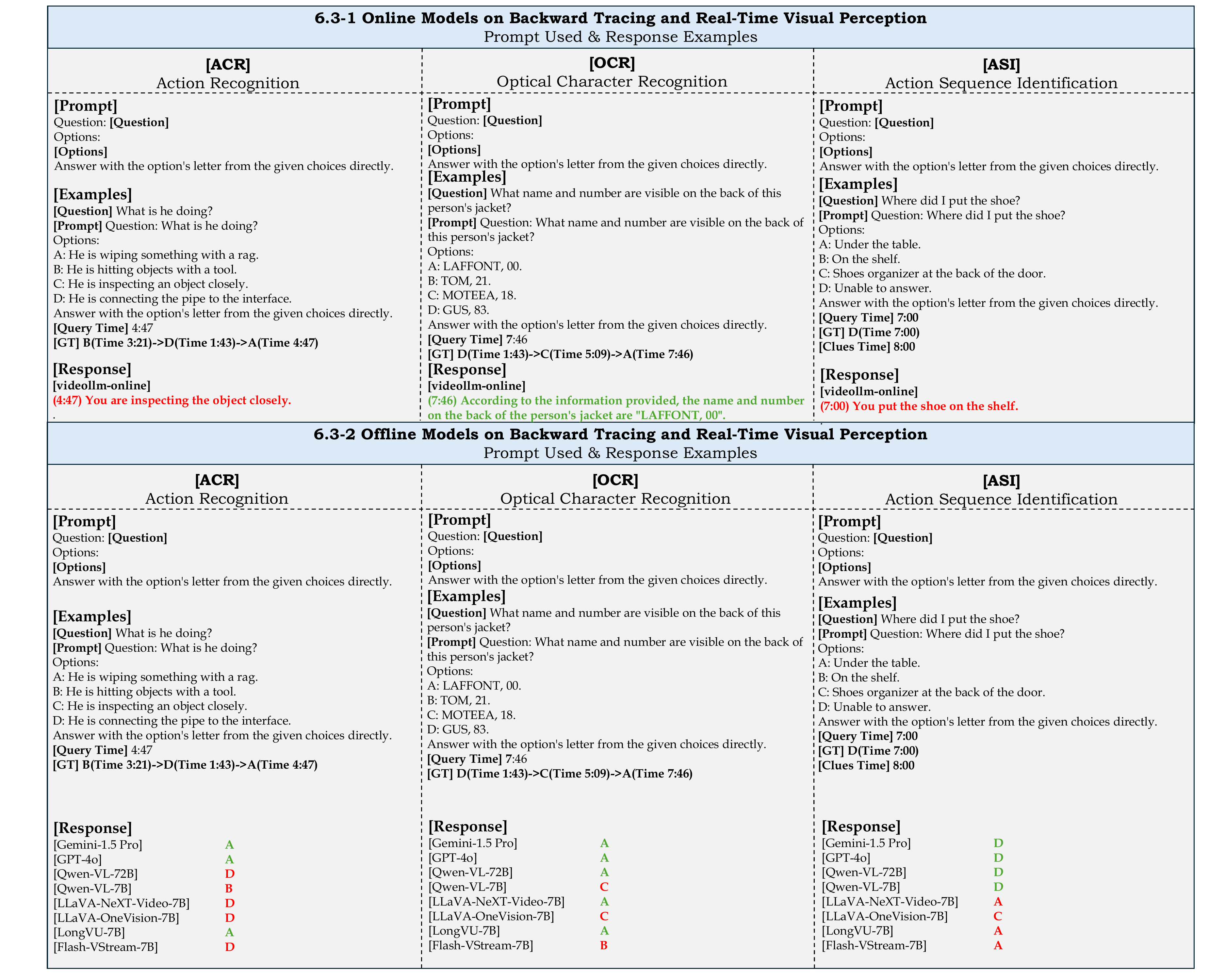}
    \vspace{-0.5cm}
    \caption{
    \textbf{Prompts used for Online(up) and Offline(down) Models on \textit{Real-Time Visual Perception} and Response Examples.} 
    Three tasks including [ACR], [OCR], and [ASI] are included as demonstrations. Our benchmarks involve a large ratio of questions, whose answers shift over time, which means that models can hardly figure out the answer by randomly selecting frames from original videos.
    }
    \label{fig:case_study_realtime}
    \vspace{-0.3cm}
\end{figure*}

\section{More Details of Evaluation}
\subsection{Evaluation for Online Models on Forward Active Responding}
As no existing online models can satisfy the demand imposed by our original designs, we choose not to cover this part in our main paper. We introduce an effective evaluation metric tailored for each task consisting of two different dimensions. \par
\textbf{Guidance for evaluation metrics design.}
\begin{itemize}
    \item \textbf{Accuracy-Based.} The model's responses should, first of all, be correct without misleading information. We judge the effectiveness of the answer given by the model, and simply average all of them to give the accuracy.
    \item \textbf{Score-Based.} Based on the accuracy-based evaluation, we encourage the response to be both accurate and timely and therefore devise a scoring metric.
\end{itemize}
\par
\textbf{Details of evaluation metrics.}
Given the user's queries $Q_{t_i}$ at time $t_i$, the referred events $E_{j}$ (such as a specific step of a tutorial procedure) with the time interval from $t_j$ to $t_j'$, the appropriate response $A_m$ at time $t_m$, the model's responses $R_{m'}$ at time $t_m'$, the evaluation function $F(R_{m'}, A_m)$, which directly compare the models' responses against the right ones, the evaluation metrics of different models are formally given as follows.
\begin{enumerate}
    \item \textbf{[REC]}
    In this task, the query is only made at a certain time before one complete repetition event happens. In our benchmark, the query is made at the start of the video, i.e. only $Q_{0}$ is made.
    \begin{itemize}
        \item \textbf{Accuracy-Based.} $$ Acc = \frac{\sum_{i=1}^{N}F(R_{m'}, A_m)}{N}$$

        \item \textbf{Score-Based.} $$ Score = \sum_{i=1}^{N}e^{i\cdot p_1}\cdot F(R_{m'}, A_m) \cdot 2^{-(m'-m)\cdot p_2} $$
    \end{itemize}
    where $F(R_{m'}, A_m) = [A_m == R_{m'}]$, which gives $1$ if the model's response is the same as the answer, and gives $0$ otherwise. $p_1$ and $p_2$ are parameters to balance the weight. In our evaluation, they are set to $0.2$ and $0.05$ respectively.
    
    \item \textbf{[SSR]} In this task, a query like \textit{Illustrate me on how to make a sandwich according to the video} is made before the start of the procedure. Akin to [REC], the query is only made at the start of the video, i.e. only $Q_0$ is made.
    \begin{itemize}
        \item \textbf{Accuracy-Based.} $$ Acc = \frac{\sum_{i=1}^{N}F(R_{m'}, A_m)}{N} $$
        \item \textbf{Score-Based.} $$ Score = \sum_{i=1}^{N}F(R_{m'}, A_m) \cdot 2^{-(m'-m)\cdot p} $$
    \end{itemize}
    where we leverage GPT-4o to give $F(R_{m'}, A_m)$, measuring the effectiveness of $R_{m'}$ given the reference answer $A_m$ and relevant visual content. $p$ is set to 0.5 to balance weight in our evaluation.
    
    \item \textbf{[CRR]} In this setting, queries are made before every $A_m$, i.e. $range(i)==range(m)$.
    \begin{itemize}
        \item \textbf{Accuracy-Based.} $$ Acc = \frac{\sum_{i=1}^{N}F(R_{m'}, A_m)}{N} $$
        \item \textbf{Score-Based.} $$ Score = \sum_{i=1}^{N}F(R_{m'}, A_m) \cdot 2^{-(m'-m)\cdot p} $$
    \end{itemize}
    where we leverage GPT-4o to give $F(R_{m'}, A_m)$, measuring the effectiveness of $R_{m'}$ given the reference answer $A_m$ and relevant visual content. $p$ is set to 0.5 to balance weight in our evaluation.
    
\end{enumerate}
\par

\textbf{Prompt Design.} To adapt to the online scenarios, we constructed streaming mode prompts with accurate timestamps and also deleted the complicated instructional statement compared to \ref{sec:offline-FAR}. Prompts and examples of models' responses are shown in \ref{fig:case_study_forward}.

\subsection{Prompt Design for Offline Models on Forward Active Responding}\label{sec:offline-FAR}
The \textit{Forward Active Responding} task is intrinsically inappropriate for offline models, as these models only support queries about existing video contents and can not receive additional visual frames after the query is made. However, considering the superiority of offline models against existing online models, we design a multiple-triggering evaluation pipeline and prompt offline models to decide whether the current time is appropriate for answering the user's query. Formally, given the user's query $Q_{t_0}$ at $t_0$, we leverage offline models to decide at $t_{i}, i\geq 1; t_{i}>t_0$ whether video contents from $t_0$ to $t_{i}$ offer sufficient clues. Specifically, for each of the tasks under the \textit{Forward Active Responding} setting, instructional prompts and examples of models' \cite{videollm-online}\cite{zhang2024flashvstreammemorybasedrealtimeunderstanding} responses are shown in Fig. \ref{fig:case_study_forward}. \par

\subsection{Prompt Design for Models on Backward Tracing and Real-Time Visual Perception}
We use the clip from the beginning to the query time to query models. Prompts and examples of models' responses are shown in Fig. \ref{fig:case_study_realtime}.

\section{More Details of Benchmark Construction}

\subsection{Human-annotated QA Generation}
We leverage meticulous human labor for part of the QA generation.\par
\textbf{Real-Time Visual Perception.} For tasks, including [STU], [OJR], and [ATR], we invite volunteers to propose candidate questions in supplement to our Video-LLMs-based automatic generation pipeline. This procedure is designed to alleviate possible bias and increase diversity. Specifically, we provide our volunteers with the following guidelines:
\begin{itemize}
    \item Watch the video and decide whether this candidate is appropriate for constructing questions that can be classified into the above three types.
    \item Selected appropriate moments for problem construction. Consider whether the moment contains: 1. Obvious spatial relationships between several objects; 2. Interested objects, such as something that appears in the moment, and so on; 3. Objects with unusual attributes, e.g. green fire, smooth woods.
    \item Construct options for the questions. Ensure that 1. Options should be relevant to the visual content; 2. Incorrect options should bring misleading information from the visual content; 3. Options should be as close in length as possible.
\end{itemize}
\par
\textbf{Clue Reveal Responding.} For our novel [CRR] task, we find it difficult to construct satisfactory question proposals by straightly prompting Video-LLMs with original video content as reference or LLMs with the provided scripts and subtitles as reference. So we recruit volunteers to propose queries and corresponding answers. Our guidelines for volunteers are as follows:
\begin{itemize}
    \item Find scenes with apparent discontinuity. For example, character A performs a certain action at query time $Q_i$. However, the action's complete process or outcome is not immediately shown during query time. 
    \item Continue watching the video, find clues for your query, and annotate the clues revealing time as $A_i$. 
    \item Try to provide concise timestamps, let $A_i$ be the time when enough visual information has just been revealed. 
\end{itemize}

\section{Additional Dataset Analysis}
\subsection{Task and Sample Distribution}
\begin{figure}[t]
    \centering
    \includegraphics[width=0.8\columnwidth]{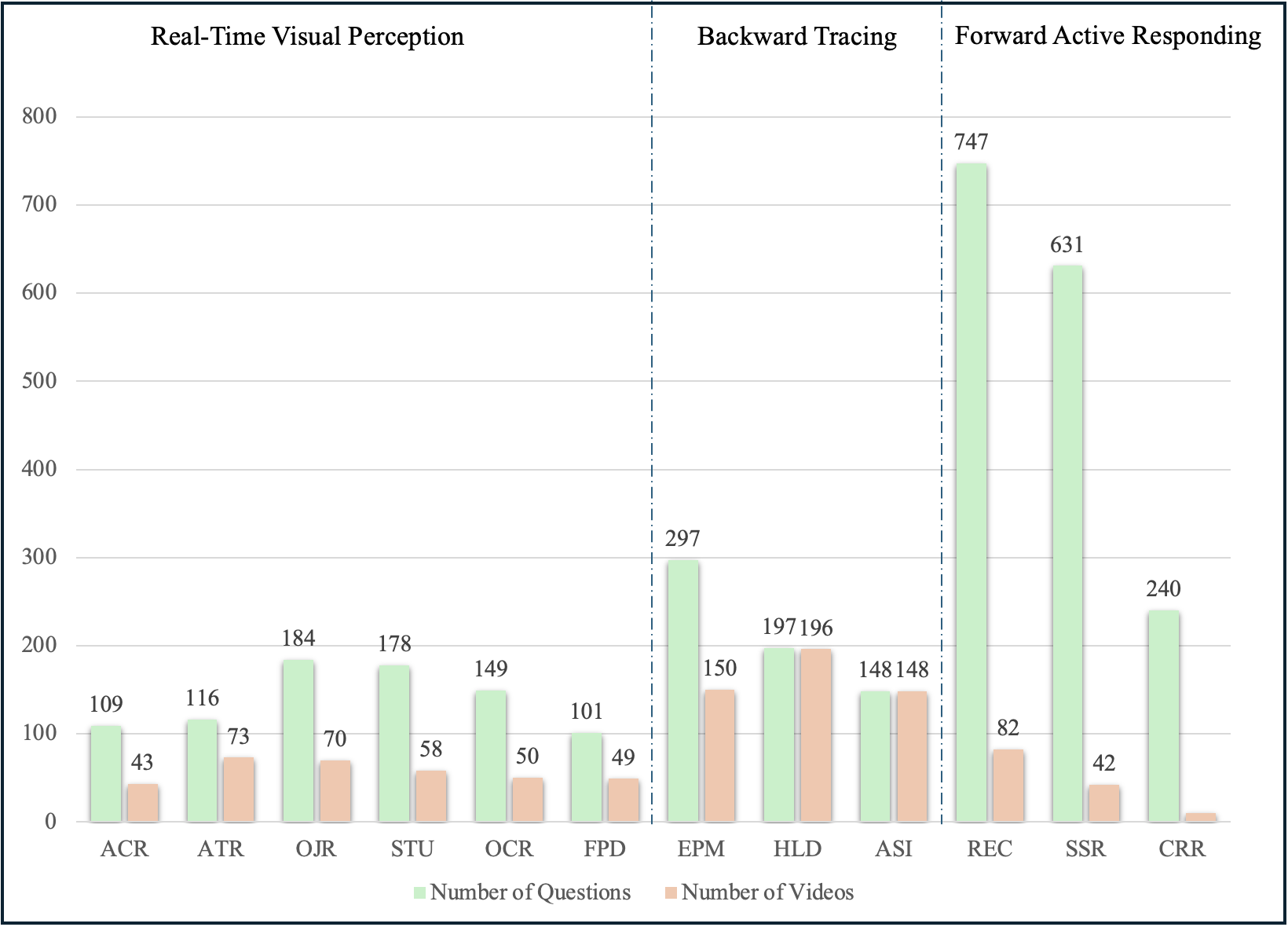}
    \vspace{-10pt}
    \caption{
    \textbf{Distribution of questions and video in OVO-Bench.}}
    \label{fig:data_num}
    \vspace{-15pt}
\end{figure}

Fig. \ref{fig:data_num} illustrates the distribution of questions and videos in OVO-Bench across the twelve tasks listed in Fig. \ref{fig:pdfpart}.
\subsection{Query Timestamps and Video Duration}
\begin{figure}[t]
    \centering
    \includegraphics[width=0.8\columnwidth]{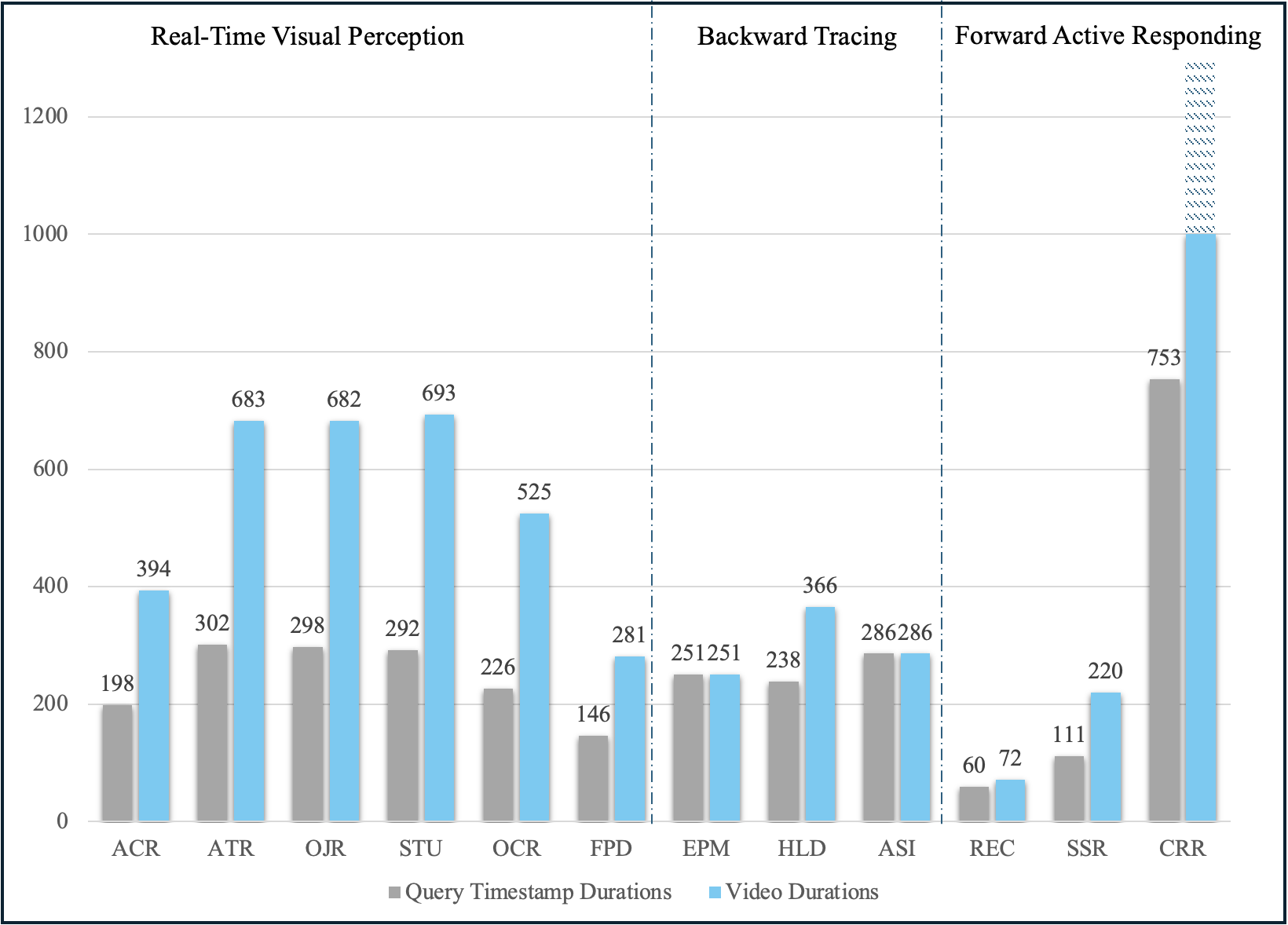}
    \vspace{-10pt}
    \caption{ \textbf{Distribution of averaged query timestamps and video duration (in seconds) in OVO-Bench.} Specifically, the averaged video duration in CRR is ~6,857 seconds.}
    \label{fig:data_duration}
    \vspace{-15pt}
\end{figure}

Fig. \ref{fig:data_duration} illustrates the distribution of averaged query timestamps and video duration in OVO-Bench across the twelve tasks listed in Fig. \ref{fig:pdfpart}.

\section{Limitations}
 While we have tried hard to cover a wide range of reasonable video domains and QA generation methods, the scarcity of existing datasets with annotations that fit requirements, the unsatisfactory results of automatic QA generation, and the high human annotation cost, hinder diversity and can cause potential bias. 
 \par
 \textbf{Offline Models for Online Video Understanding.} As implied in our analysis \ref{fig:case_study_forward}, offline models usually perform random guesses in the forward active responding scenarios, making our evaluation unfair. For example, a model that always outputs \textit{"Yes"} can still achieve a score above zero in our evaluation. Moreover, the absence of online models with satisfactory performance, makes our benchmarks more suitable for future advancements. We hope our intensive work and intuitive ideas can guide the development of video understanding models toward real-world online video understanding.

\section{Licenses}
The annotations of our OVO-Bench are provided to the community under CC BY-NC-SA 4.0 license. By downloading our dataset from our website or other sources, the user agrees to adhere to the terms of CC BY-NC-SA 4.0 and licenses of the source datasets. Download links are provided for our self-crawled YouTube videos. Licenses of the source datasets are listed in \ref{table:license}

\begin{table}[ht]
    \centering
    \resizebox{0.30\textwidth}{!}{
        \begin{tabular}{l|l}
            \toprule
            \textbf{Dataset} & \textbf{License}  \\
            \midrule
            QAEgo4D \cite{barmann2022did} & N/A  \\
            OpenEQA \cite{majumdar2023openeqa} & MIT License \\
            STAR \cite{wu2024star} & Apache License 2.0  \\
            HiREST \cite{zala2023hierarchical} & MIT License  \\
            YouCook2 \cite{zhou2018towards} & MIT License  \\
            CrossTask \cite{zhukov2019cross} & BSD 3-Clause License  \\
            COIN \cite{tang2019coin} & Research Purpose Only \\
            Ego4D \cite{grauman2022ego4d} & MIT License \\
            THUMOS'14 \cite{jiang2014thumos} & Research Purpose Only \\
            THUMOS'15 \cite{gorban2015thumos} & Research Purpose Only  \\
            Perception Test \cite{patraucean2024perception} & CC BY 4.0  \\
            MovieNet \cite{huang2020movienet} & N/A \\
            E.T.Bench \cite{liu2024etbench} & CC BY 4.0 \\
            \bottomrule
        \end{tabular}
    }
    \caption{License of source datasets in OVO-Bench.}
    \label{table:license}
\end{table}

\section{Data Examples}
We provide more examples extracted from our benchmark. We try to cover different video categories in every task to offer a holistic overview of OVO-Bench.

\newpage


\section*{Supplementary material}

# [EPM] Episodic Memory

## QA-Ego4D

**Clue Time: 5:22**
Question: Where was the kitchen paper towel before I picked it?
Options: **A. top of the kitchen sink**; B. under the sink; C. on the counter; D. on the stove

**Clue Time: 3:44**
Question: What did I pick from the fridge?
Options: A. bread; B. milk; **C. vegetable**; D. water

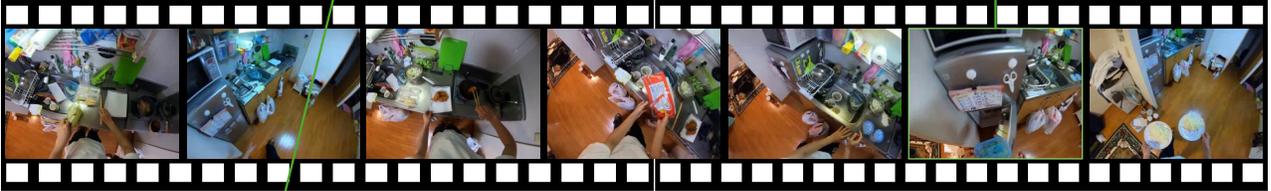
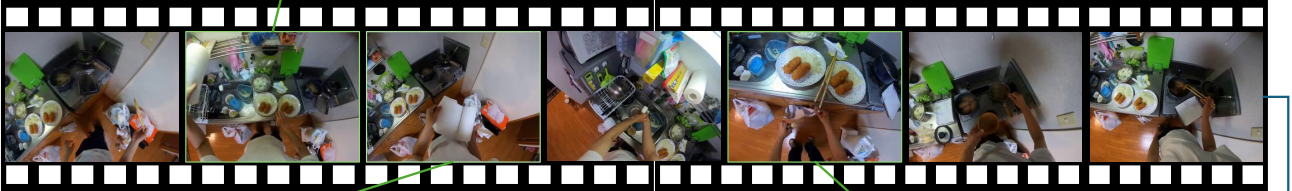

**Clue Time: 5:25**
Question: How many rolls of paper towel did I cut?
Options: A. two; **B. one**; C. three; D. four

**Clue Time: 6:36**
Question: Did I leave the drawer open?
Options: A. yes; **B. no**

Query At the End

**Clue Time: 1:59**
Question: Where was the paper towel before I picked it?
Options: A. on the counter; B. on the cupboard; **C. on the table**; D. in the dish washer

**Clue Time: 3:05**
Question: What drawer did I pull?
Options: **A. dish washer**; B. microwave; C. cupboard; D. fridge

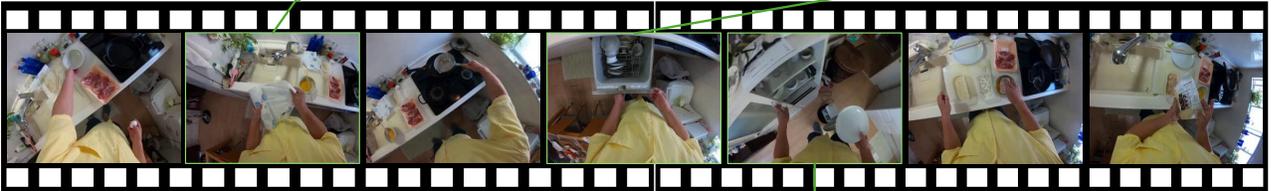
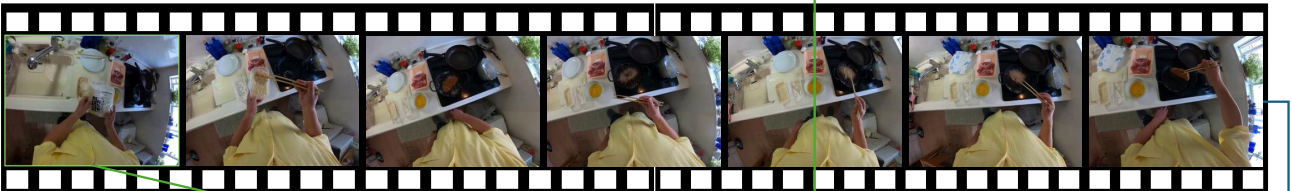

**Clue Time: 4:09**
Question: What did I pour in the container?
Options: A. flour; **B. bread crumbs**; C. salt; D. pepper

**Clue Time: 3:13**
Question: Where was the plate before I picked it?
Options: A. on the counter; **B. on the cupboard**; C. on the table; D. in the dish washer

Query At the End

## Open-EQA

**Clue Time: 0:43**
Question: What is the gold object on the nightstand?
Options: A. A painting; B. A mirror; **C. A nightlamp**; D. A vase

**Clue Time: 0:50**
Question: Is this home on the first floor?
Options: A. Yes, it's on the first floor; **B. No, it's on the second floor**;

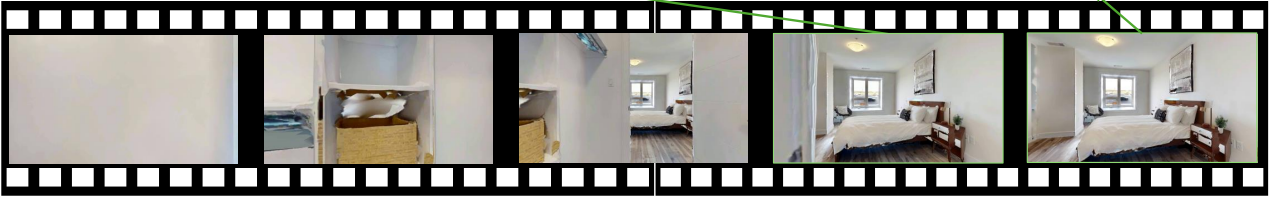
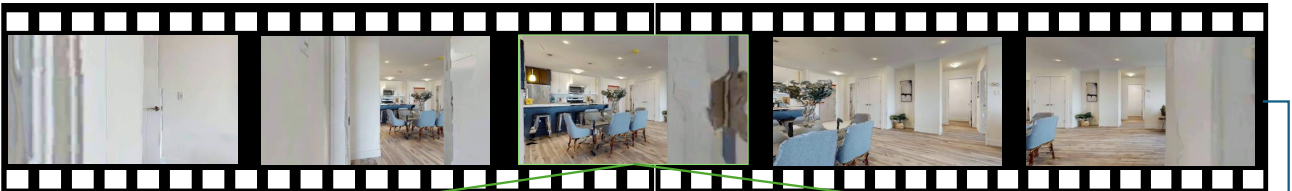

**Clue Time: 1:10**
Question: Where can I sit and eat if I don't want to use the dining table?
Options: **A. Use the kitchen bar counter**; B. Use the floor in the hallway; C. Use the bed in the bedroom; D. Use the couch in the living room

**Clue Time: 1:10**
Question: What color is the smoke detector?
Options: A. White; **B. Yellow or off-white**; C. Black; D. Blue

Query At the End

# [HLD] Hallucination Detection

**Query Time: 6:11**
Question: what did I put in the black dustbin?
Options: A. empty water bottles; **B. Unable to answer**;
C. old newspapers; D. food scraps

**Query Time: 6:44**
Question: Where did I put the vacuum cleaner head?
Options: A. closet; **B. Unable to answer**;
C. bathroom; D. kitchen

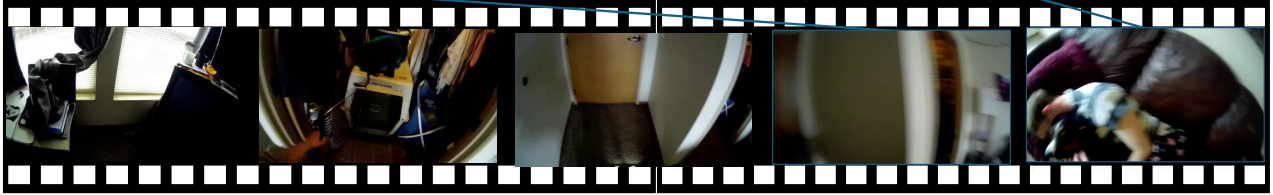
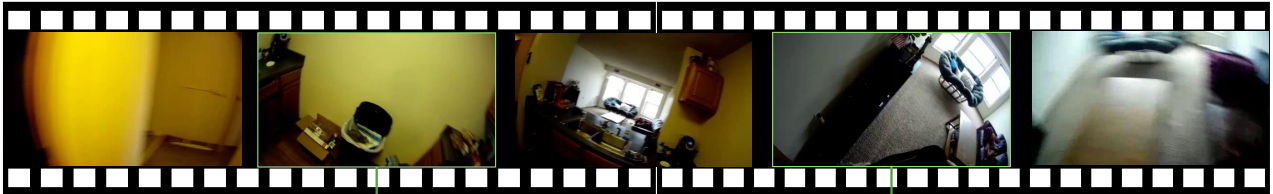

**Clue Time: 7:10**

**Clue Time: 7:30**

**Query Time: 6:48**
Question: Where were game boards?
Options: **A. Unable to answer**; B. in the shelves;
C. in the fridge; D. in the bags

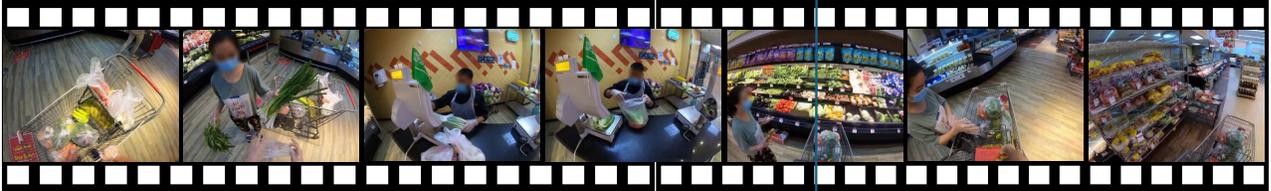
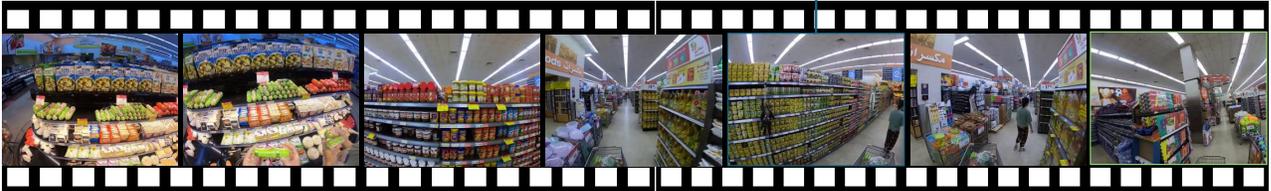

**Clue Time: 7:55**

QA-Ego4D

**Query Time: 0:35**
Question: what color is the flower in the bottom floor?
Options: A. Red; **B. Unable to answer**;
C. Blue; D. White

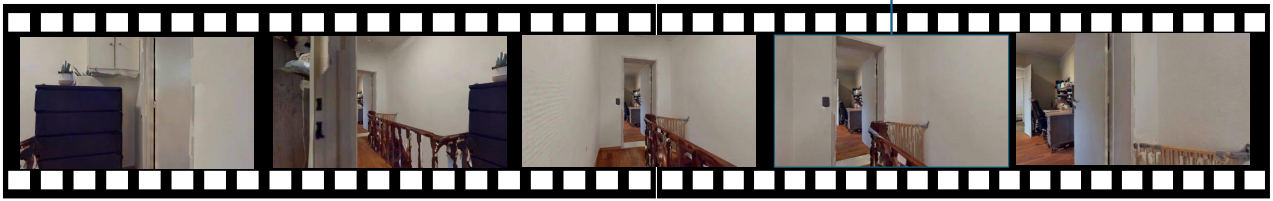
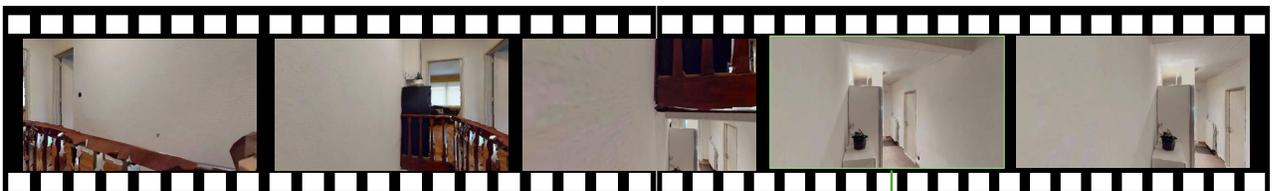

**Clue Time: 1:15**

Open-EQA

# [ASI] Action Sequential Identification

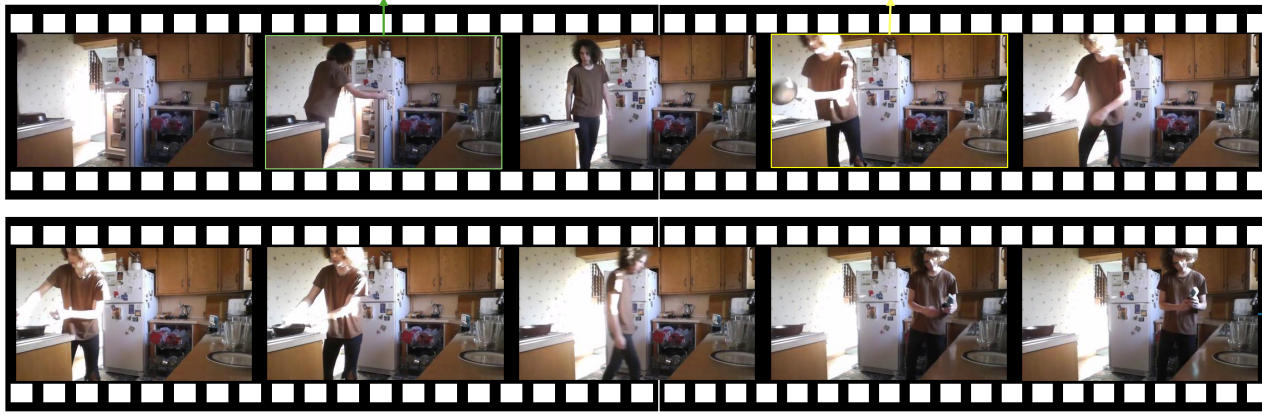

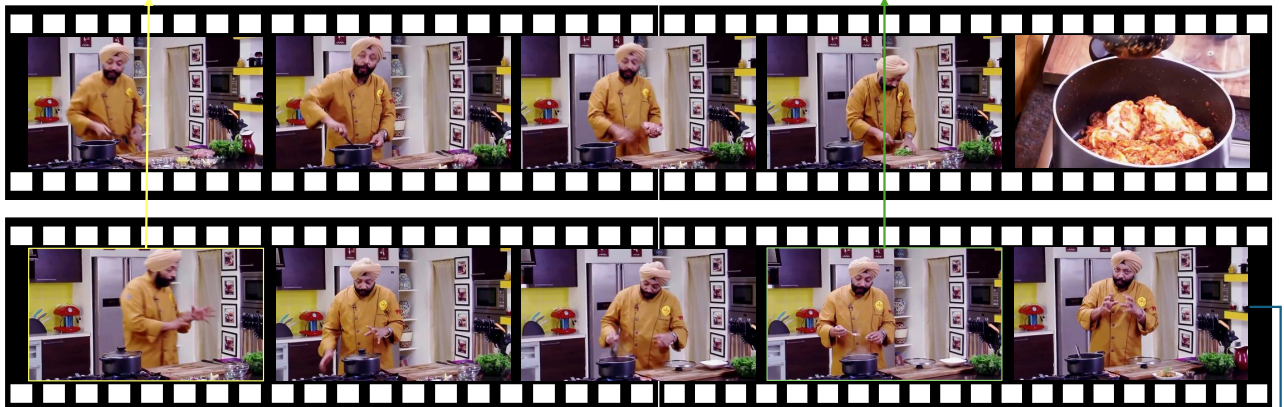

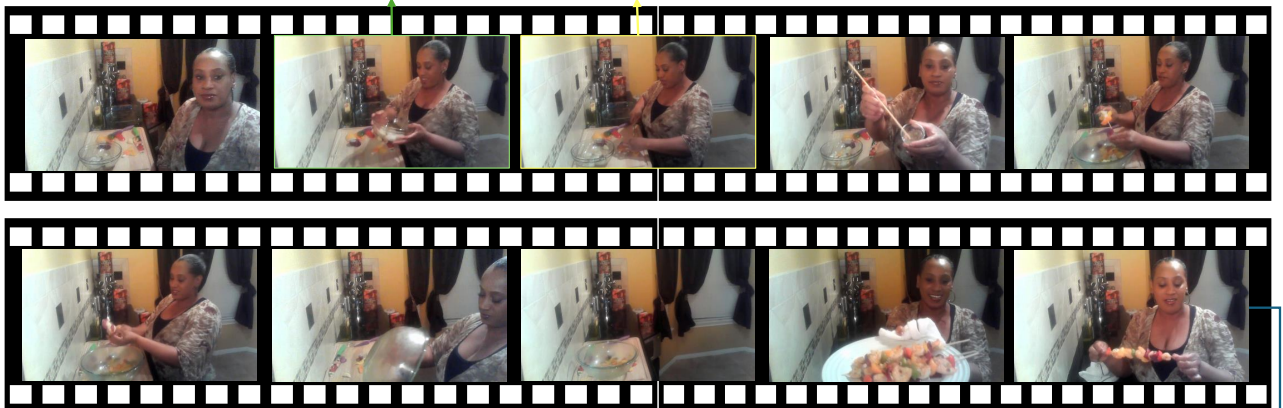

# [ASI] Action Sequential Identification

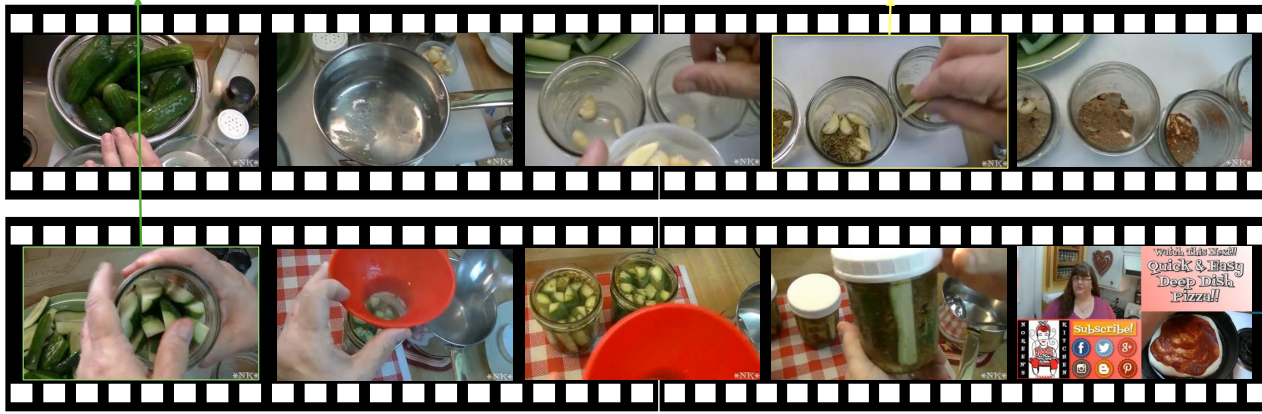

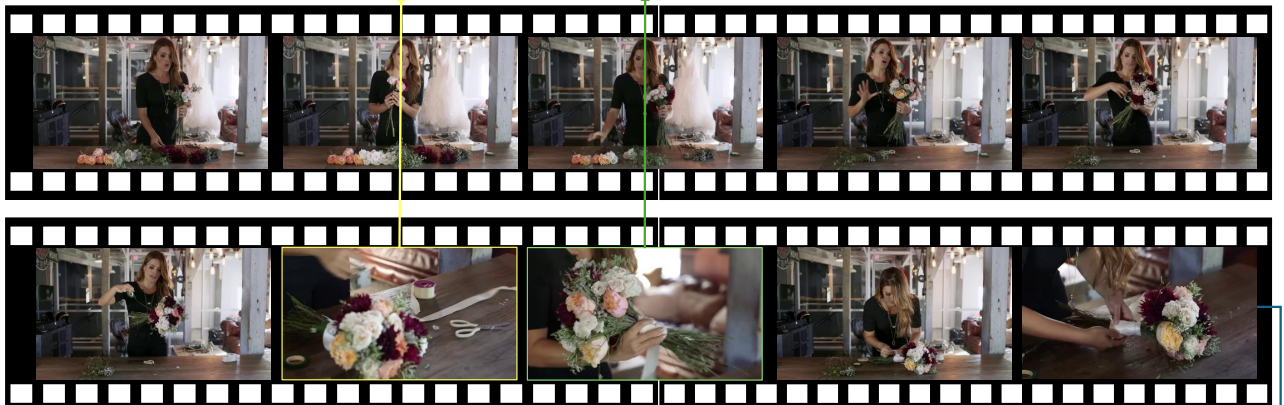

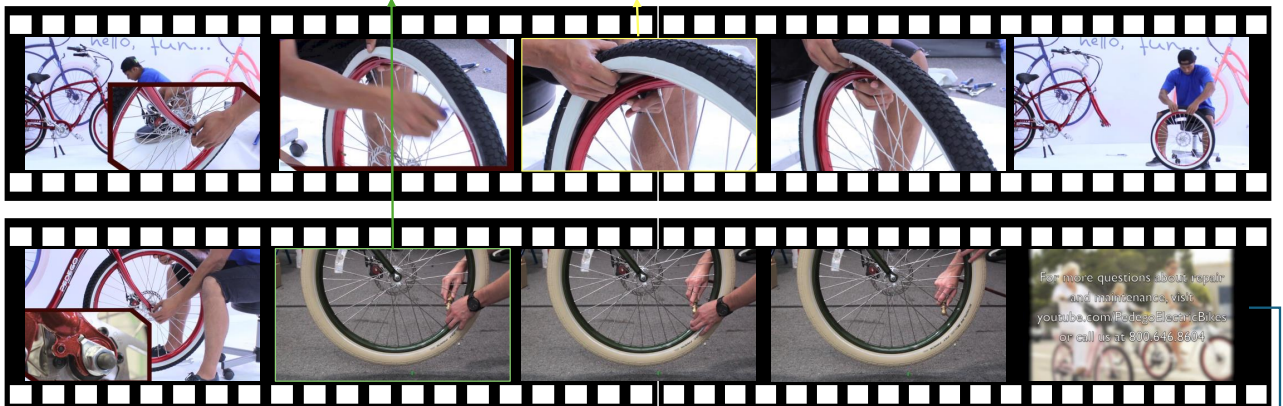

# [STU] Spatial Understanding

**Query Time/Clue Time: 1:27**
Question: What is the relative position of the person to the car ?
Options: A. The person is standing at the front of the car.;
        B. The person is on the co-pilot side of the car.;
        **C. The person is standing beside the driver's side of the car** ;
        D. The person is behind the trunk of the car.

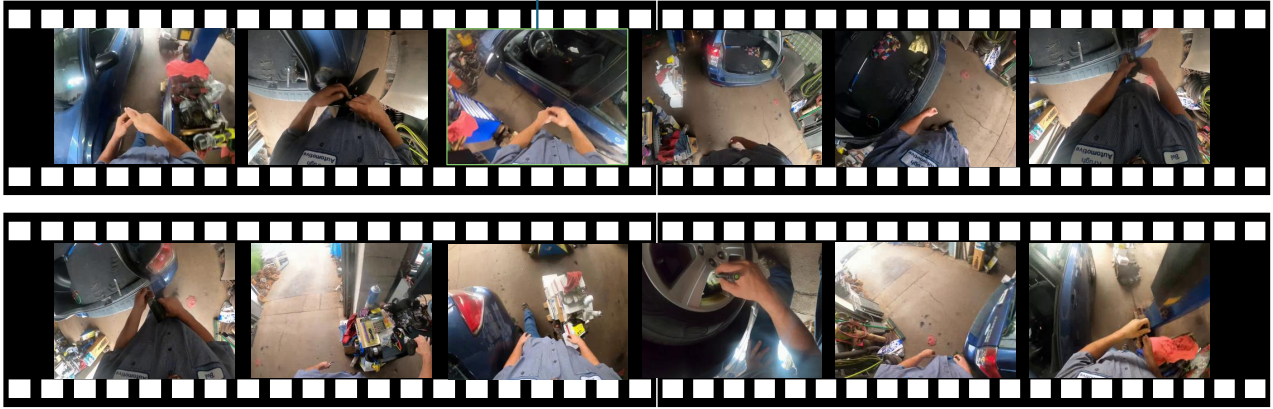

Ego4D

**Query Time/Clue Time: 1:47**
Question: Which container is located closer to the top left corner of the table?
Options: A. The blue container;
        **B. The white container**;
        C. The red container;
        D. The green container.

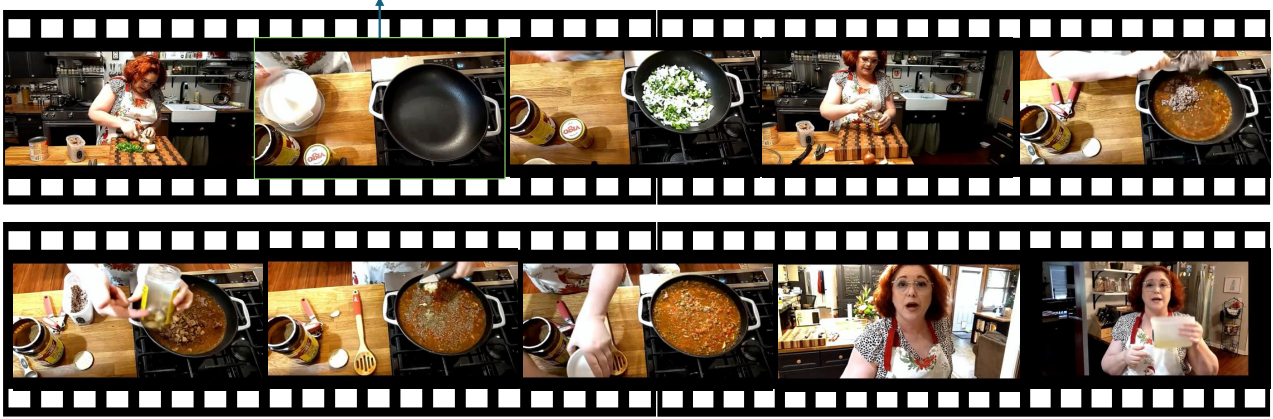

YouTube

**Query Time/Clue Time: 0:12**
Question: Which road did I take?
Options: **A. The road on the right**;
        B. The road on the left;
        C. Unable to answer.

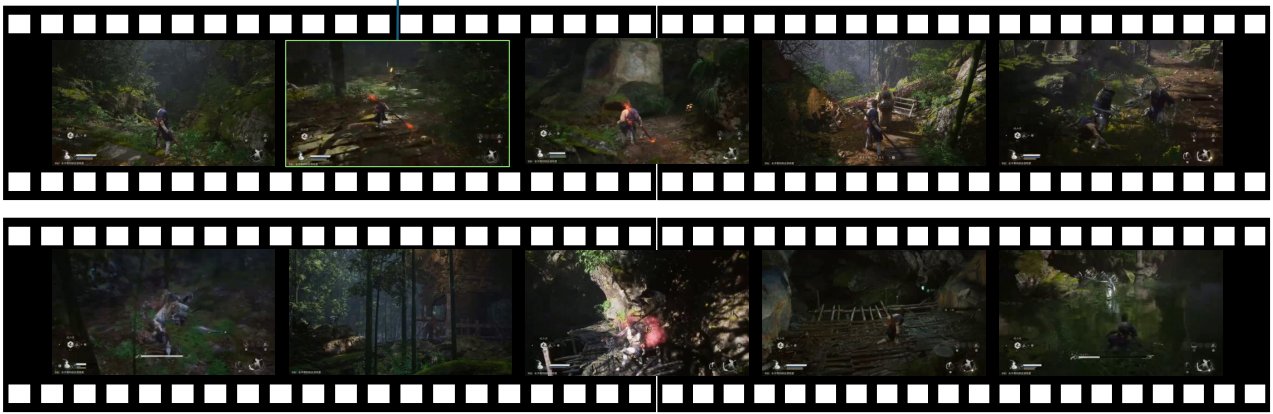

YouTube-Human Annotated

# [OJR] Object Recognition

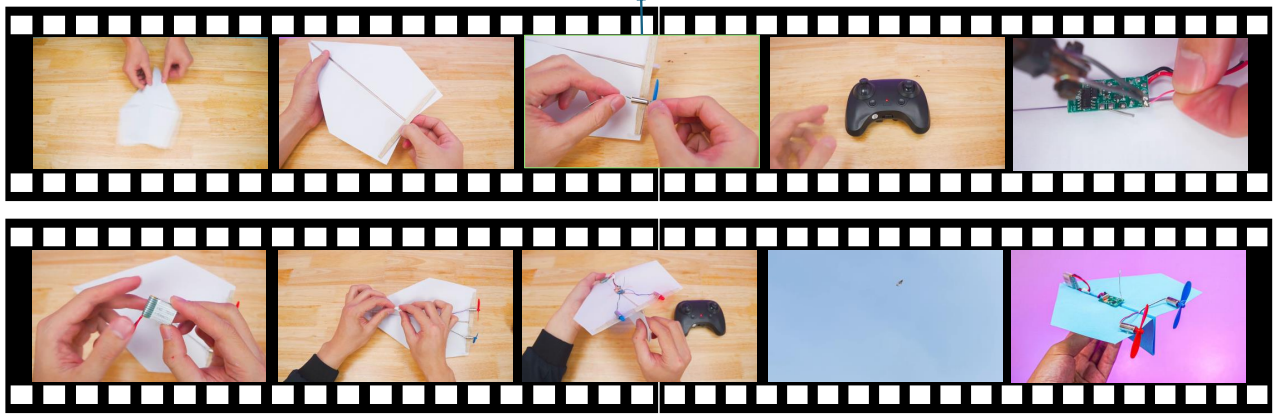

**Query Time/Clue Time: 2:08**
Question: What object is being used to construct the paper aircraft wing?
Options: A. A plastic frame.;
    **B. A wooden stick..**;
    C. A paper sheet;
    D. A metal rod.

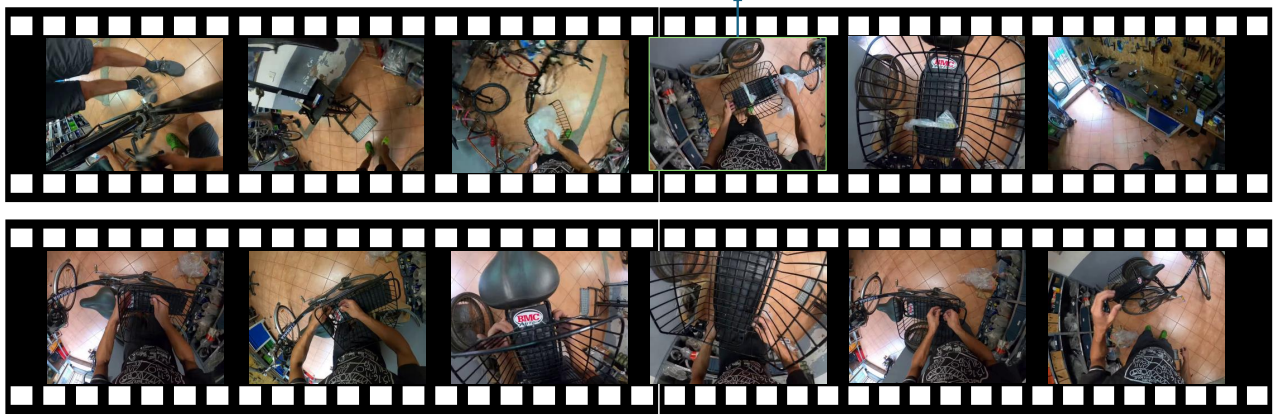

**Query Time/Clue Time: 1:27**
Question: What object is being attached to the back of it?
Options: A. A bag.;
    B. A saddlebag.;
    **C. A basket**;
    D. A pannier.

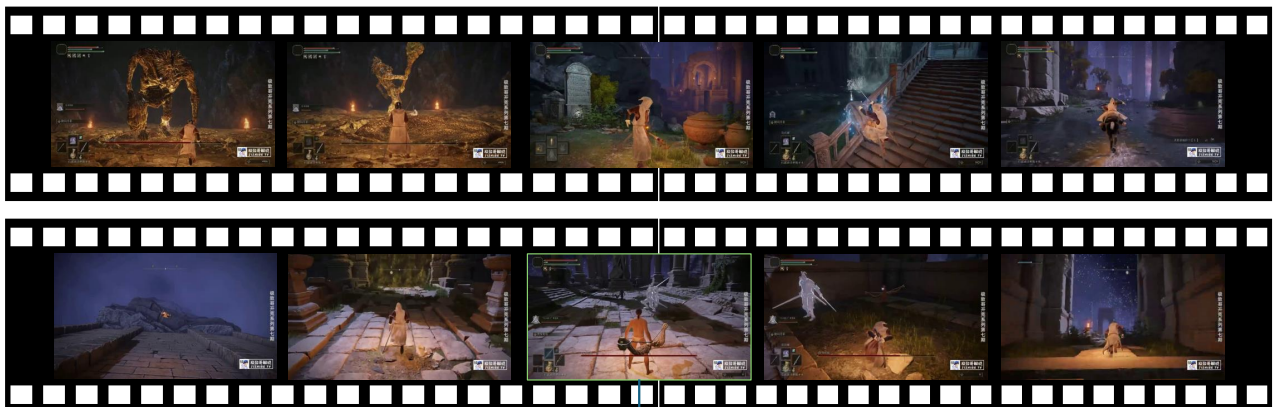

**Query Time/Clue Time: 4:28**
Question: What weapon do I have in my hand?
Options: **A. Bow**; B. Sword;
    C. Unable to answer.
    D. Axe. E. Spear

# [ATR] Attribute Recognition

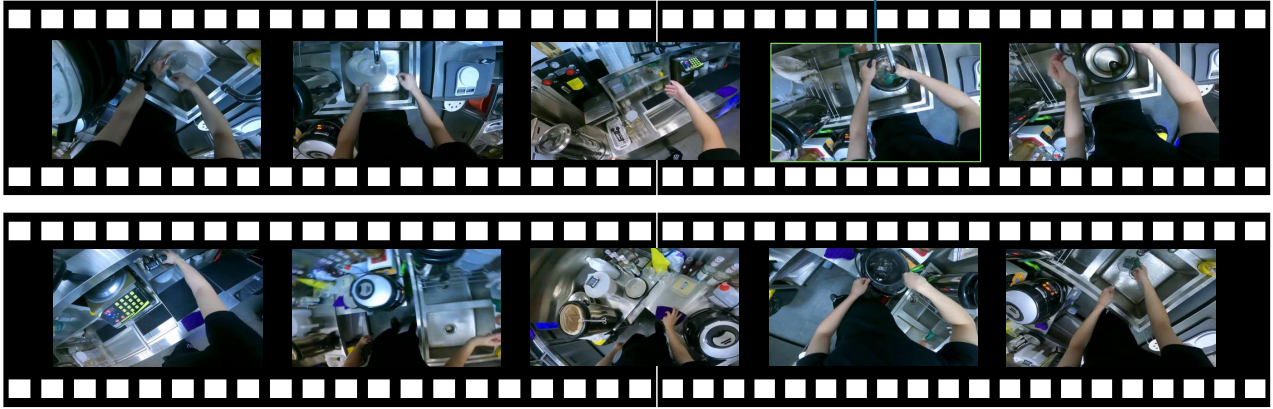

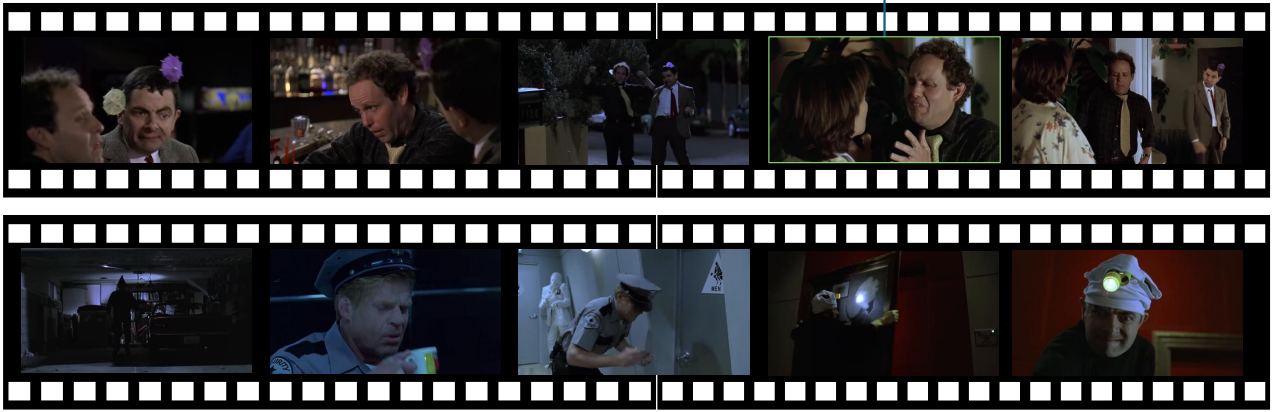

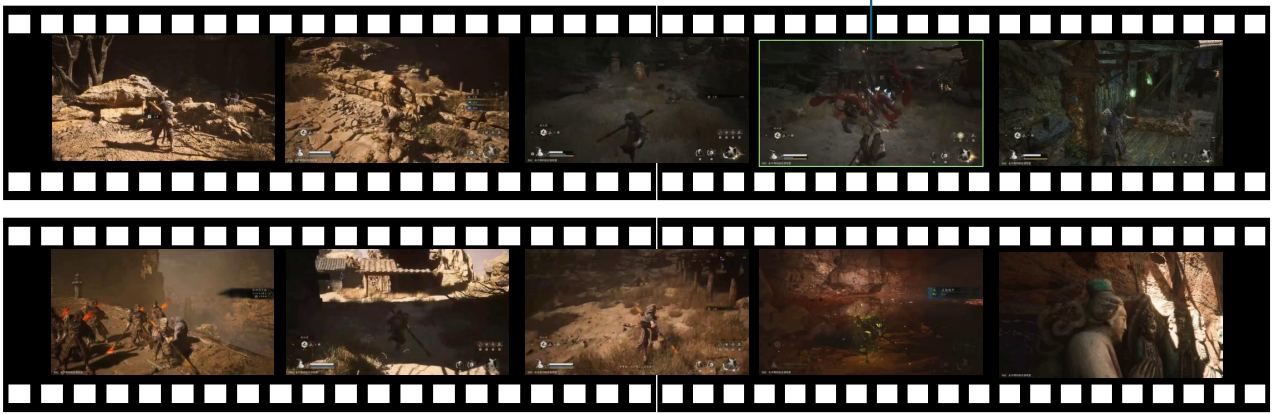

# [ACR] Action Recognition

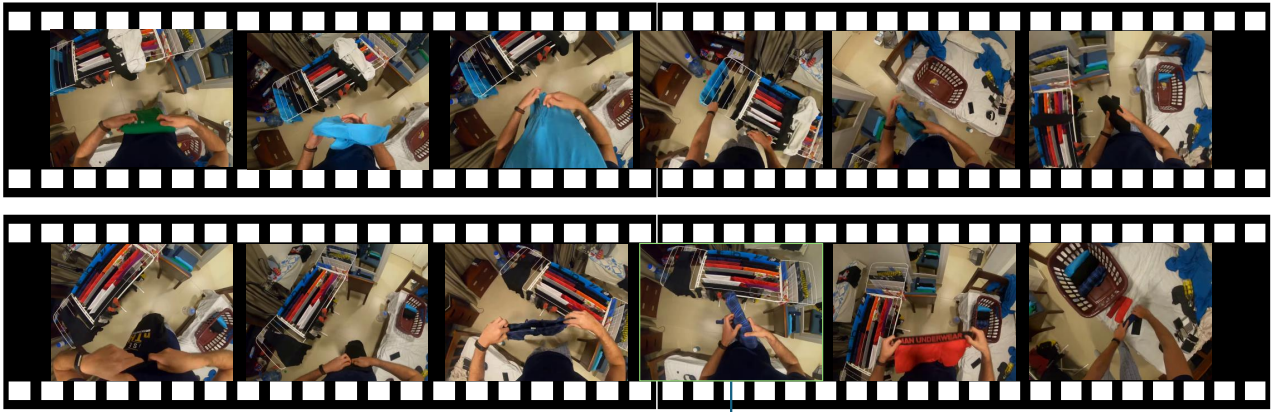

**Query Time/Clue Time: 4:18**
Question: What action is he performing with the blue checkered cloth?
Options: A. He is wiping the tabletop with the cloth.;
**B. He is folding the cloth.;**
C. He is covering a basket with the cloth;
D. He is tying the cloth around his neck.

**Query Time/Clue Time: 1:03**
Question: What is he doing?
Options: A. He is playing a musical instrument.;
B. He is taking off his clothes.;
C. He is taking a shower;
**D. He is answering the phone.**

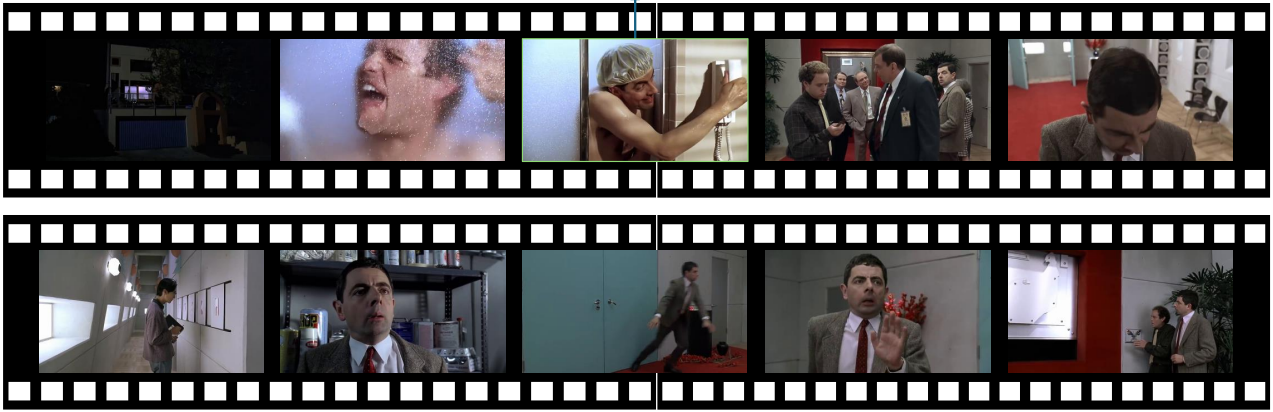

**Query Time/Clue Time: 2:37**
Question: What is she doing with the chicken in her hands?
Options: A. She is placing the chicken into a pot of water.;
**B. She is putting the chicken down.;**
C. She is holding the chicken while preparing to cook it.;
D. She is cutting the chicken into smaller pieces..

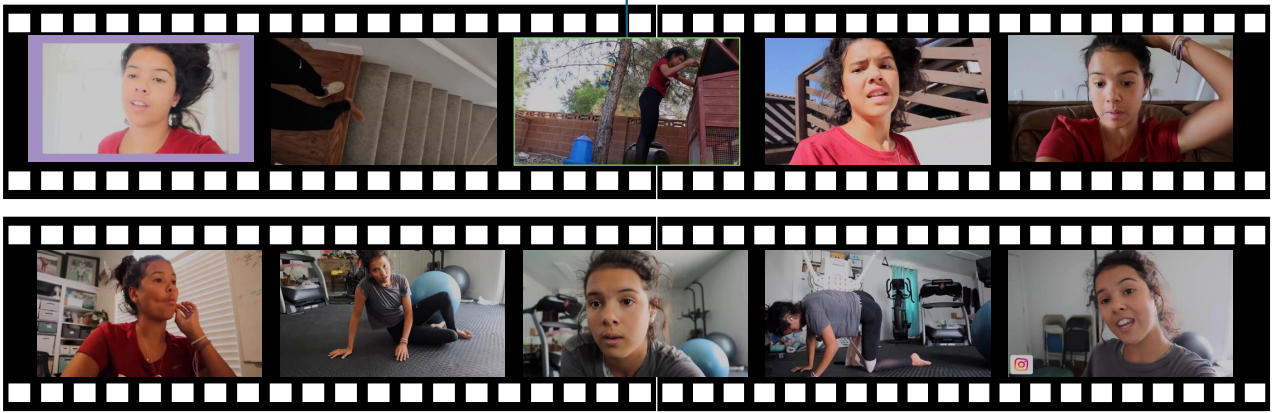

## [OCR] Optical Character Recognition

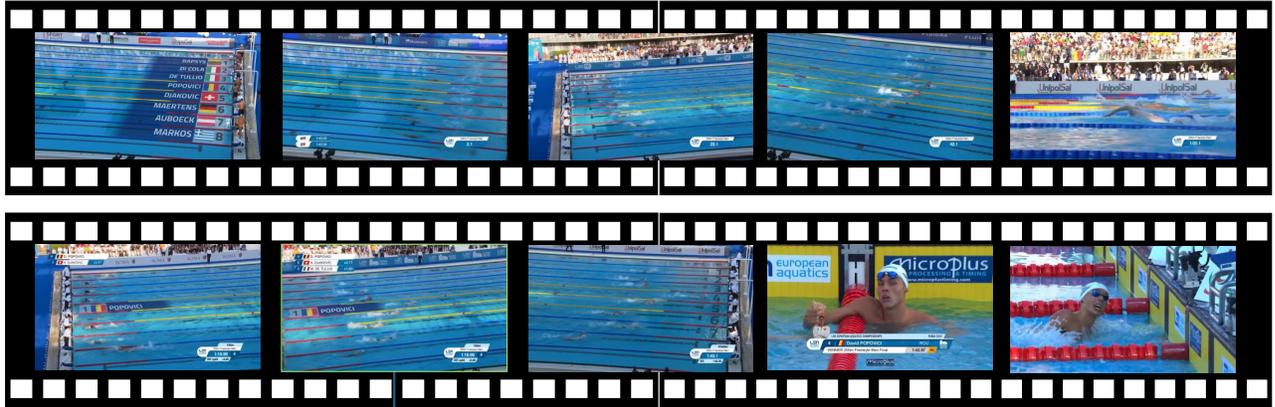

**Query Time/Clue Time: 2:16**
Question: What is the leading player's time at the 150m mark?
Options: **A. 1:16.96** B. 1:15.89;
C. 1:14.32. D. 1.19.01

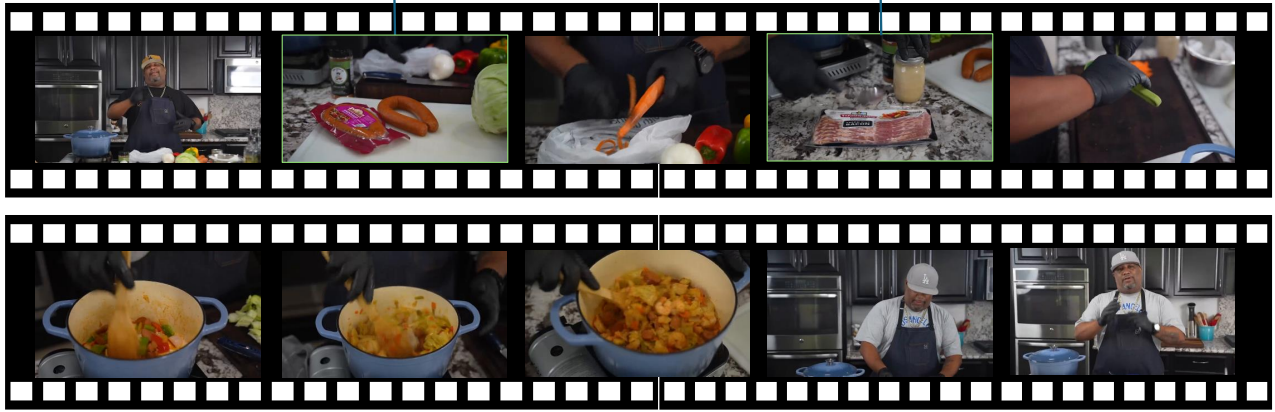

**Query Time/Clue Time: 0:23**
Question: What is the text on the package?
Options: **A. ANDOUILLE.**;
B. THICK CUT HAM.;
C. THICK CUT PORK;
D. THICK CUT BACON.

**Query Time/Clue Time: 1:35**
Question: What is the text on the package?
Options: A. ANDOUILLE.;
B. THICK CUT HAM.;
C. THICK CUT PORK;
**D. THICK CUT BACON**.

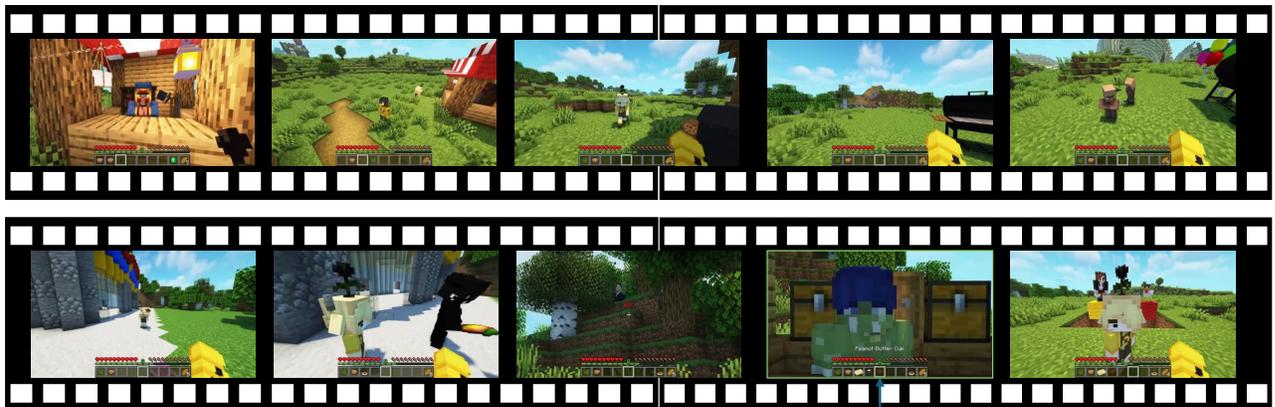

**Query Time/Clue Time: 10:12**
Question: What text is displayed now?
Options: A. Milk Bucket.;
B. Butter;
**C. Peanut Butter Cup.**;
D. Peppermint Swirl.

YouTube

# [FPD] Future Prediction

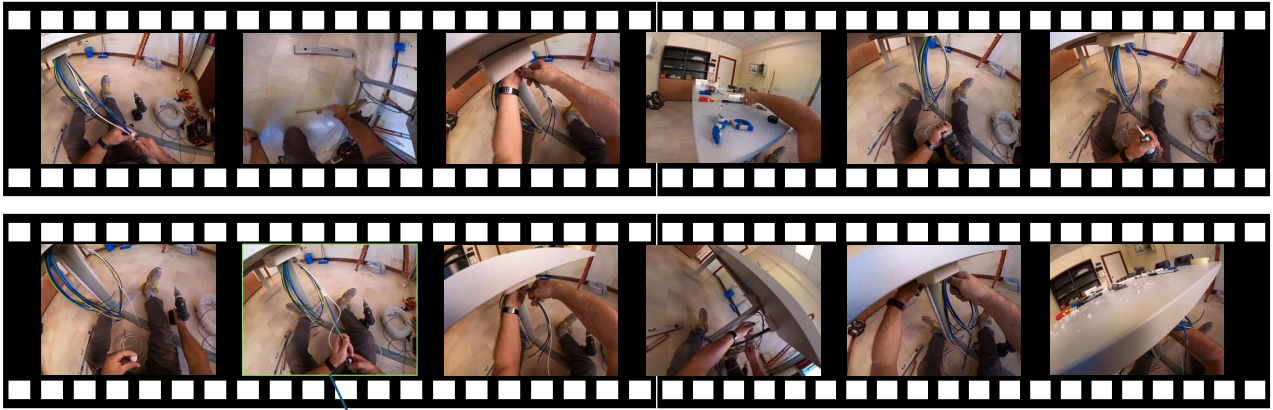

**Query Time/Clue Time: 1:45**
Question: What is the person preparing to manipulate?
Options: **A. The person is about to handle or manipulate the wire**;
　　　　B. The person is about to handle or manipulate the plastic tubing;
　　　　C. The person is about to handle or manipulate the circuit board;
　　　　D. The person is about to handle or manipulate the metal sheet

**Query Time/Clue Time: 0:59**
Question: What action is this person preparing to take ?
Options: **A. The person is about to turn on the faucet**;
　　　　B. The person is preparing to start the car engine;
　　　　C. The person is about to open the refrigerator door;
　　　　D. The person is reaching for the light switch.

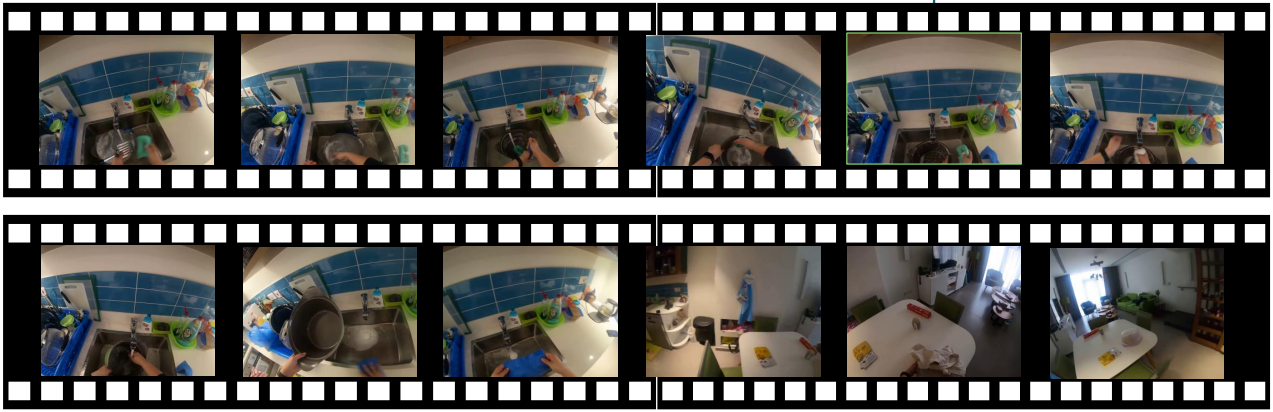

**Query Time/Clue Time: 0:59**
Question: What is this person about to do?
Options: A. The person is about to press a button.;
　　　　B. The person is about to grab a book;
　　　　**C. The person is about to open the drawer**;
　　　　D. The person is about to adjust the lamp.

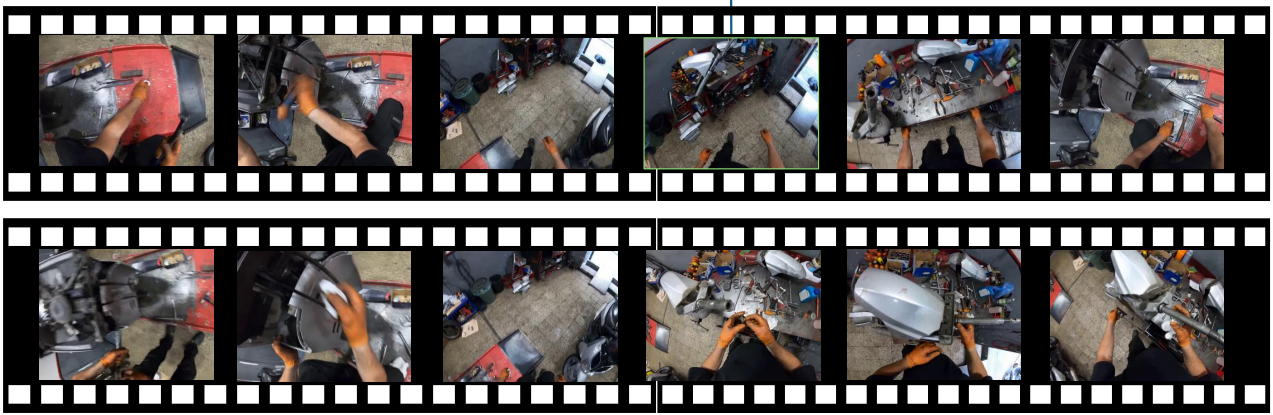

Ego4D

# [REC] Repetition Event Count

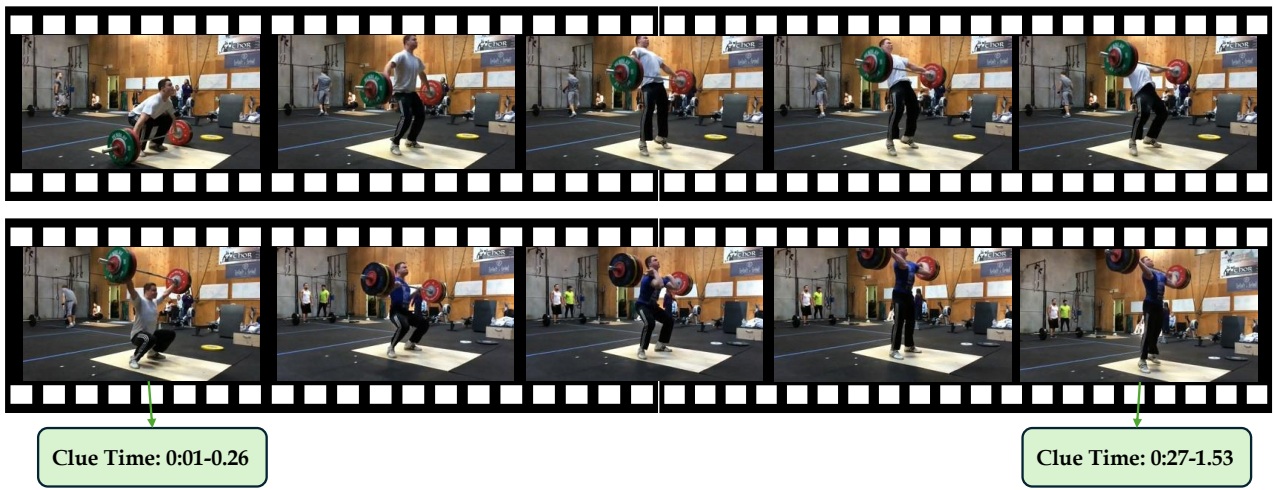

**Thumos 14/15**

Query Time: 0.26/1.53/End of the video
Question: How many times did they clean and jerk?
Answers: 1/2/2

Clue Time: 0:01-0.26
Clue Time: 0:27-1.53

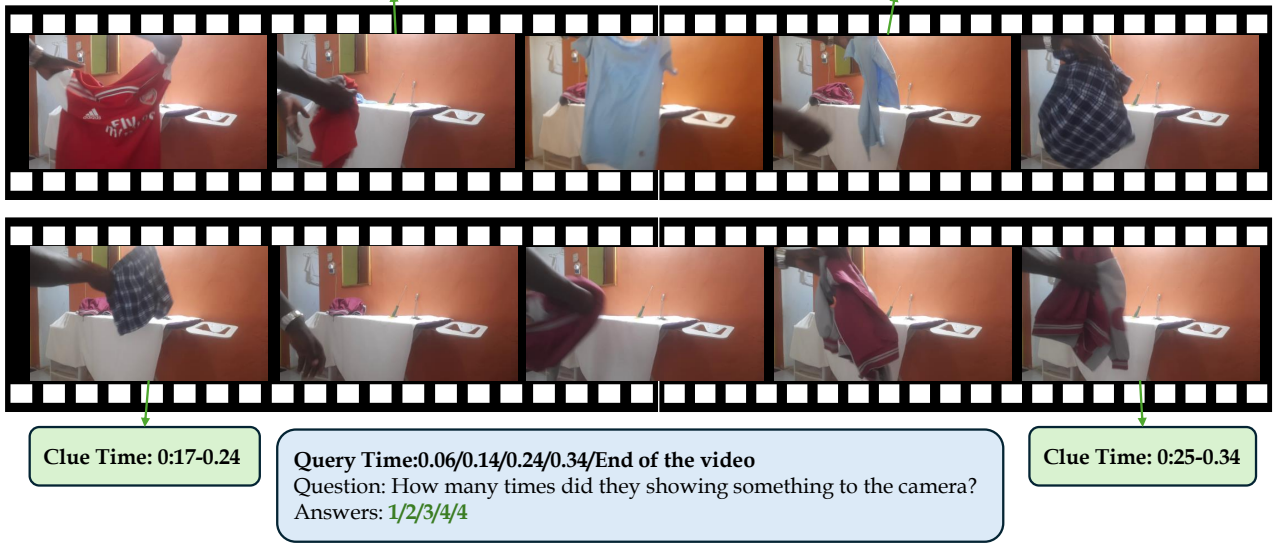

**Perception Test**

Clue Time: 0:00-0.06
Clue Time: 0:07-0.14
Clue Time: 0:17-0.24
Clue Time: 0:25-0.34

Query Time: 0.06/0.14/0.24/0.34/End of the video
Question: How many times did they showing something to the camera?
Answers: 1/2/3/4/4

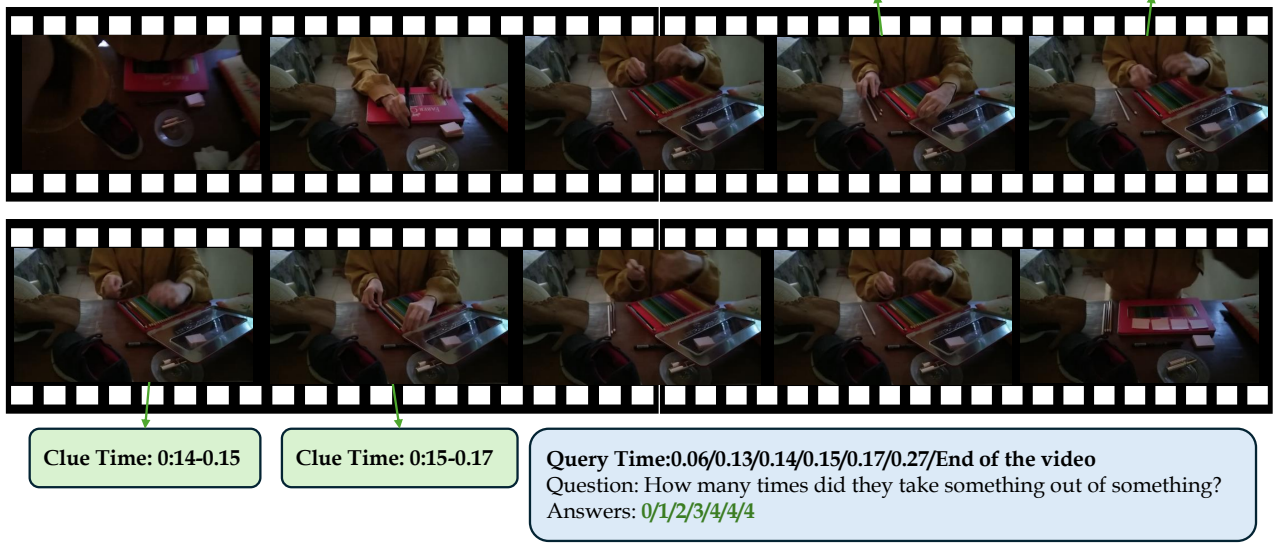

Clue Time: 0:11-0.13
Clue Time: 0:13-0.14
Clue Time: 0:14-0.15
Clue Time: 0:15-0.17

Query Time: 0.06/0.13/0.14/0.15/0.17/0.27/End of the video
Question: How many times did they take something out of something?
Answers: 0/1/2/3/4/4/4

# [SSR] Sequential Steps Recognition

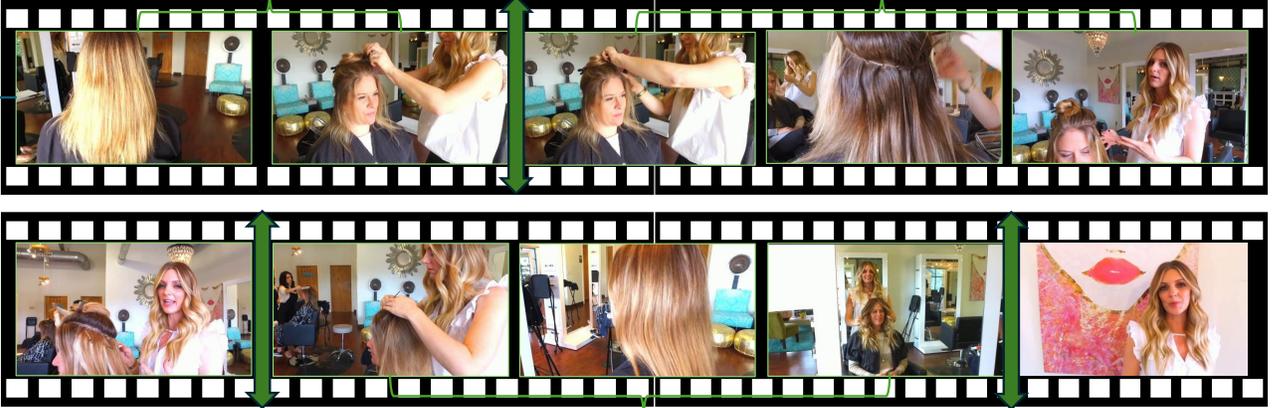

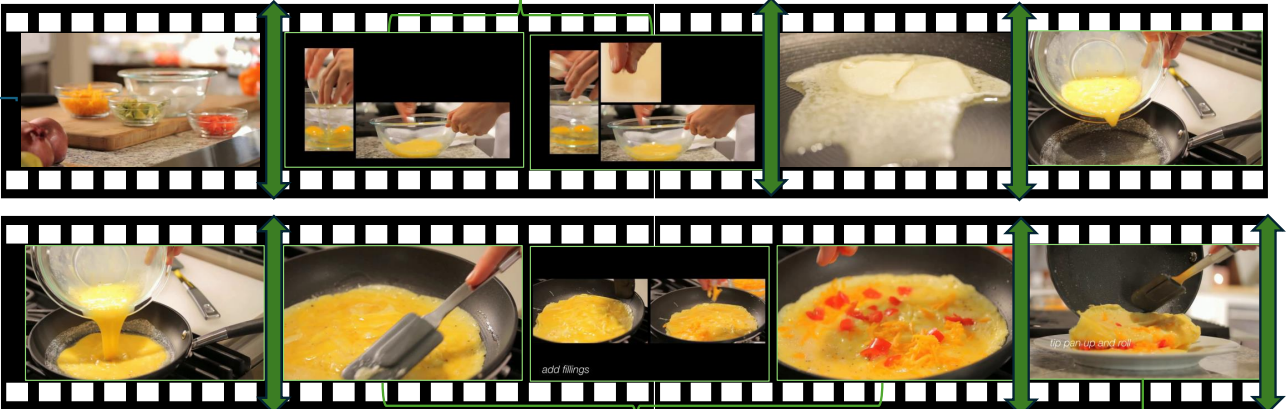

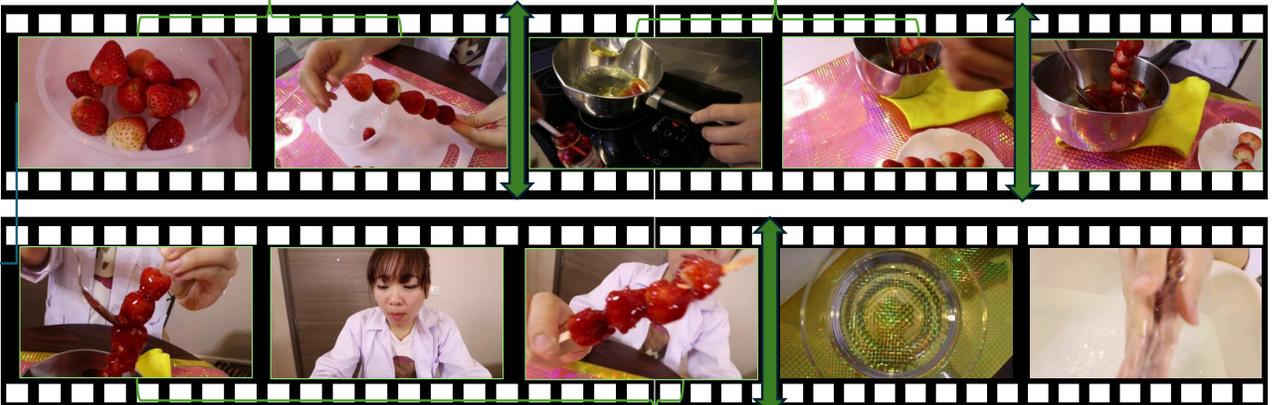

# [CRR] Clues Reveal Responding

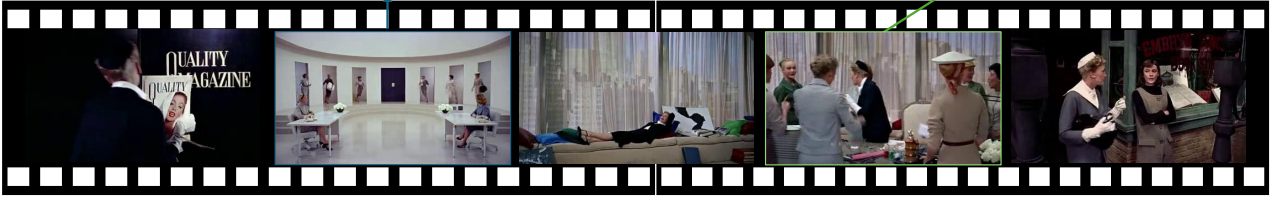

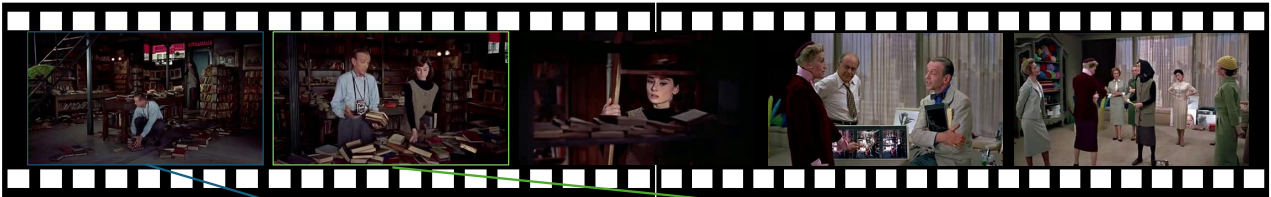

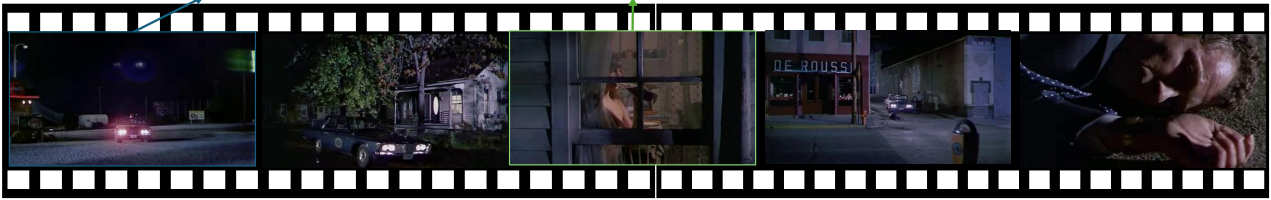

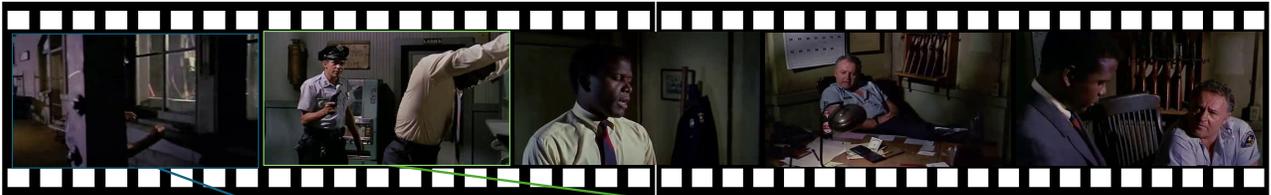

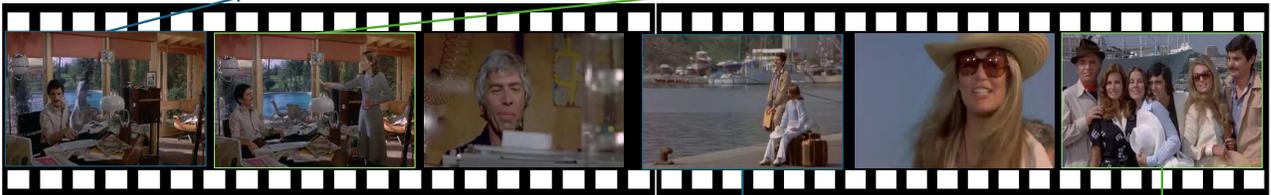

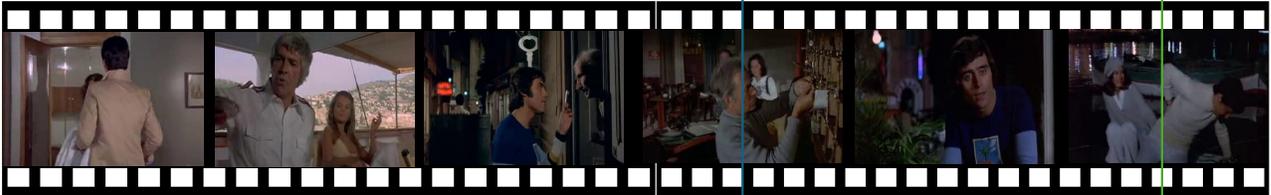



\newpage
{
    \small
    \bibliographystyle{ieeenat_fullname}
    \bibliography{main}
}

\end{document}